\documentclass[10pt,twocolumn,letterpaper]{article}

\usepackage{style}

\definecolor{paperblue}{rgb}{0.21,0.49,0.74}
\usepackage[pagebackref=false,breaklinks,colorlinks,allcolors=paperblue]{hyperref}

\usepackage{amsmath}
\usepackage{amssymb}
\usepackage{booktabs}
\usepackage{bbm}
\usepackage{color}
\usepackage{colortbl}
\usepackage{enumitem}
\usepackage{graphicx}
\usepackage{makecell}
\usepackage{multicol}
\usepackage{multirow}
\usepackage{subcaption}
\usepackage{adjustbox}
\usepackage{titletoc}
\usepackage[toc,page]{appendix}
\usepackage[most]{tcolorbox}
\usepackage{fontawesome}
\usepackage{FiraSans}

\definecolor{event_green}{RGB}{0,180,139}
\definecolor{frame_red}{RGB}{239,99,75}
\definecolor{fusion_blue}{RGB}{99,113,250}
\definecolor{planning_gold}{RGB}{218,165,32}

\definecolor{ev_1}{RGB}{0,122,204}
\definecolor{ev_2}{RGB}{0,136,196}
\definecolor{ev_3}{RGB}{0,150,183}
\definecolor{ev_4}{RGB}{0,165,170}
\definecolor{ev_5}{RGB}{0,180,160}
\definecolor{ev_6}{RGB}{0,188,150}
\definecolor{ev_7}{RGB}{0,194,140}
\definecolor{ev_8}{RGB}{0,200,130}
\definecolor{ev_9}{RGB}{0,191,165}  

\newcommand{\eventdrive}{\textbf{\textsl{\textcolor{ev_1}{E}\textcolor{ev_2}{v}\textcolor{ev_3}{e}\textcolor{ev_4}{n}\textcolor{ev_5}{t}\textcolor{ev_6}{D}\textcolor{ev_7}{r}\textcolor{ev_8}{i}\textcolor{ev_9}{v}\textcolor{ev_9}{e}}\nobreak\hspace{0.07em}}\xspace}

\newcommand{\perception}{
  \includegraphics[width=0.044\linewidth]{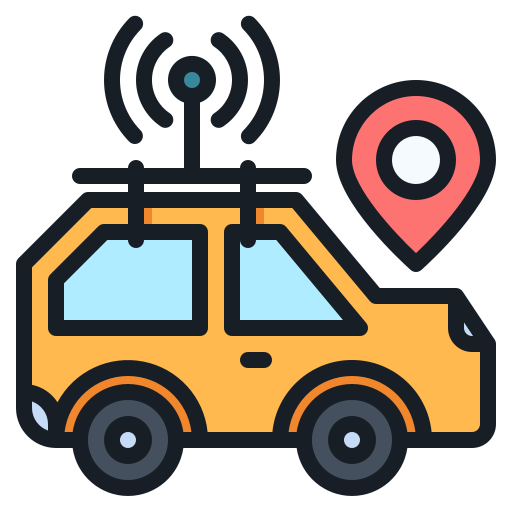}~\textbf{Perception}
}
\newcommand{\understanding}{
    \includegraphics[width=0.044\linewidth]{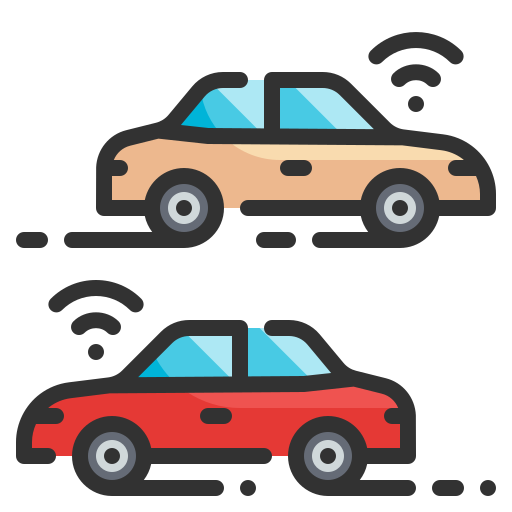}~\textbf{Understanding}
}
\newcommand{\prediction}{
    \includegraphics[width=0.044\linewidth]{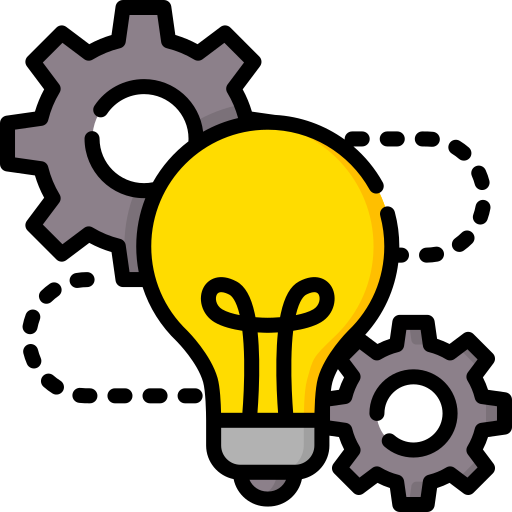}~\textbf{Prediction}
}
\newcommand{\planning}{
    \includegraphics[width=0.044\linewidth]{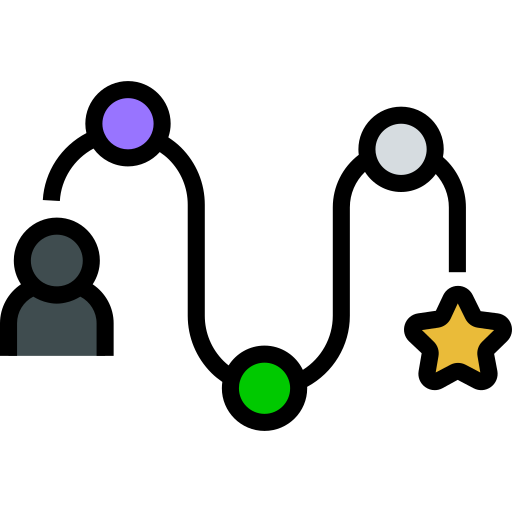}~\textbf{Planning}
}

\title{\eventdrive: Event Cameras for Vision-Language Driving Intelligence}

\author{
Dongyue Lu$^{1,6}$\quad Rong Li$^{2}$\quad Ao Liang$^{1}$\quad Lingdong Kong$^{1,3}$
\\
Wei Yin$^{4}$\quad Lai Xing Ng$^{5}$\quad Benoit R. Cottereau$^{6,7}$\quad  Camille Simon Chane$^{8}$\quad Wei Tsang Ooi$^{1,6}$
\\[1ex]
$^1$NUS\quad $^2$HKUST(GZ)\quad $^3$CNRS@CREATE\quad $^4$Horizon Robotics\quad $^5$A*STAR, I$^2$R
\\
$^6$IPAL, CNRS IRL 2955, Singapore\quad $^7$University Toulouse, CNRS, CerCo, Toulouse, France
\\
$^8$ETIS UMR 8051, CY Cergy Paris University, ENSEA, CNRS, France
\\[1ex]
~\textbf{Project Page:} \href{https://dylanorange.github.io/projects/eventdrive}{github/EventDrive}\\
~\textbf{Dataset \& Toolkit:} \href{https://huggingface.co/datasets/dylanorange/EventDrive}{huggingface/EventDrive}
}

\begin{document}

\twocolumn[{
    \renewcommand\twocolumn[1][]{#1}
    \maketitle
    \begin{center}
    \centering
    \captionsetup{type=figure} 
    \includegraphics[width=0.98\textwidth]{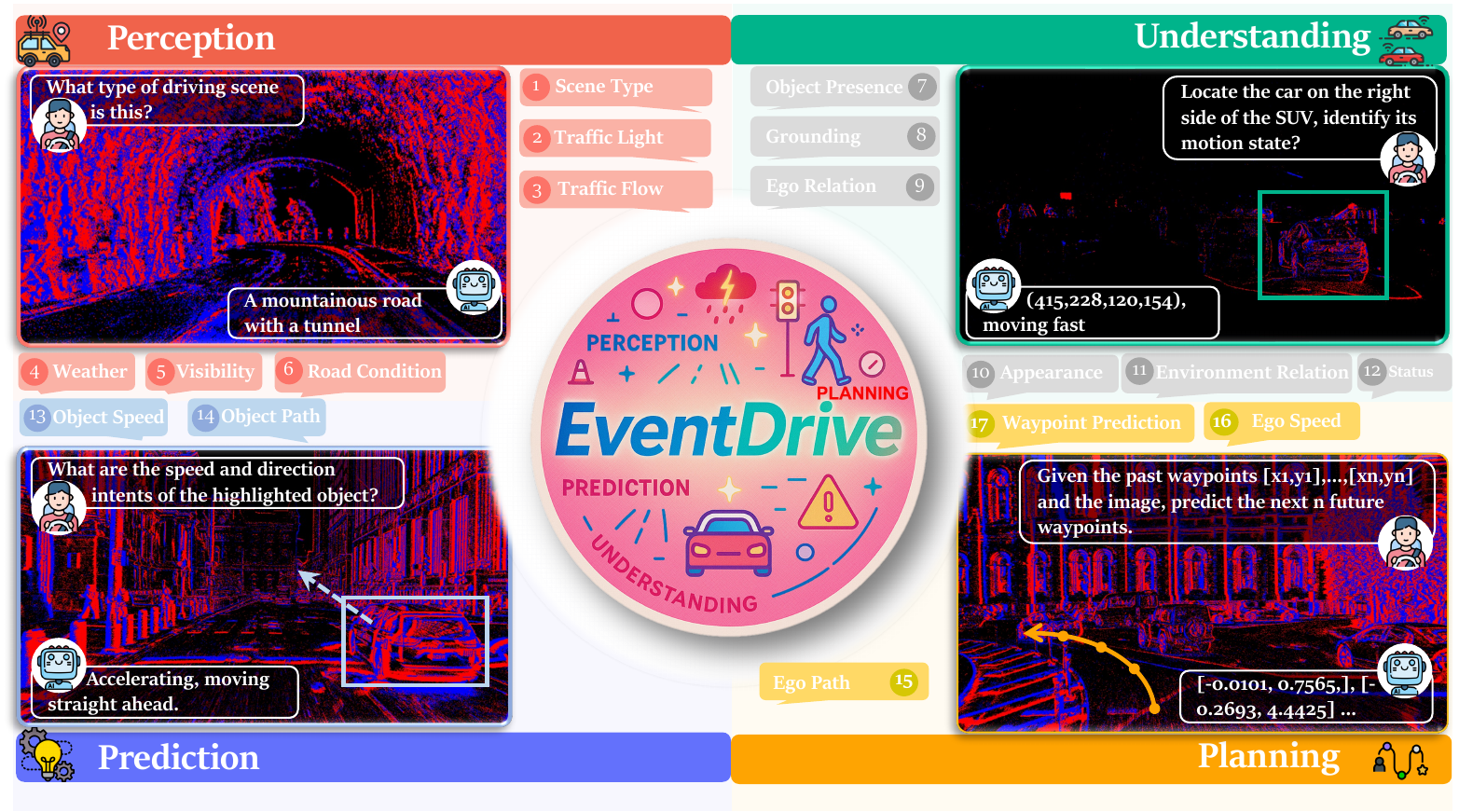}
    \caption{
    \textbf{Overview of the \eventdrive benchmark.} 
    The dataset contains $\mathbf{471k}$ event–frame–language samples across four levels of driving reasoning spanning $\mathbf{17}$ subtasks: 
    \includegraphics[width=0.016\linewidth]{icons/perception.png}~\textbf{Perception} evaluates \textit{scene-level context} such as scene type, traffic, and illumination. 
    \includegraphics[width=0.016\linewidth]{icons/understanding.png}~\textbf{Understanding} assesses \textit{object-centric semantics}, including presence, motion state, and grounding. 
    \includegraphics[width=0.016\linewidth]{icons/prediction.png}~\textbf{Prediction} infers short-horizon \textit{motion intent of surrounding agents}. 
    \includegraphics[width=0.018\linewidth]{icons/planning.png}~\textbf{Planning} estimates \textit{ego motion intent and future waypoints} from past trajectories. 
    Each task is formulated through structured, language-grounded queries, enabling unified evaluation of event–frame models across the autonomy stack.
    }
    \label{fig:teaser}
    \end{center}
}]

\begin{abstract}
Event cameras sense the world through asynchronous brightness changes with microsecond latency and high dynamic range, offering motion fidelity far beyond frame-based sensors and capturing temporal structure that conventional exposures often miss. These properties make events a powerful complement to RGB in autonomous driving, especially under blur, glare, and rapid motion, where frame-based perception can become unreliable. However, existing event-aware vision–language models remain limited to generic perception and do not reveal how event sensing contributes to reasoning and decision-making across the full driving loop. We present \eventdrive, a large-scale benchmark and model suite that unifies event streams, RGB frames, and language supervision across four core dimensions: \textbf{Perception, Understanding, Prediction, and Planning}, covering captions, structured QA, grounding, motion-state recognition, trajectory forecasting, and planning tasks. Building on this foundation, \eventdrive-VLM introduces a multi-horizon event pyramid and a temporal-horizon mixture-of-experts module to adaptively encode and fuse asynchronous and frame-based information for downstream reasoning. Comprehensive evaluation across diverse tasks shows that event streams provide substantial gains in temporal precision, motion awareness, and robustness, bringing event sensing into the center of driving intelligence.
\end{abstract}
\section{Introduction}
\label{sec:intro}

Event cameras have gained increasing attention for their ability to record dynamic scenes with microsecond temporal precision \cite{chakravarthi2024survey, gallego2022survey}. Unlike RGB sensors that integrate intensity over fixed exposure periods, event cameras report asynchronous brightness changes at the pixel level \cite{kong2025eventfly}. This design yields high dynamic range, very low latency, and natural robustness to motion blur. These sensing advantages are especially valuable in autonomous driving, where reliable perception must be maintained across fast ego motion, rapidly moving agents, and challenging illumination conditions \cite{steffen2019neuromorphic, chen2023clip2Scene}. These characteristics position event sensing as a promising complement to frame-based perception in safety-critical driving systems.

Despite this potential, events for driving research remain fragmented. Most existing efforts focus on upstream supervised tasks, \eg, detection \cite{gehrig2023rvt, li2023sodformer, lu2024flexevent}, segmentation \cite{sun2022ess, jing2024hpl-ess, kong2024openess}, or optical flow estimation \cite{zhu2018ev-flownet, zhang2022spiking}. Very few works explore high-level reasoning and decision-making that a complete driving stack requires. In contrast, the RGB community has made rapid progress toward unified vision and language models that couple perception, reasoning, and control within a single architecture \cite{hu2023planning, weng2024drive, li2025seeground, li2025_3eed, kong2025multi, survey_vla4ad}, which provides interpretability and scalable instruction-driven behavior. Early attempts to incorporate events into vision and language systems, including grounding \cite{kong2025talk2event} and caption-based event language models \cite{liu2024eventgpt, zhou2025llafea, li2025eventvl}, demonstrate encouraging multimodal interactions but remain limited to generic scenes. They do not address the reasoning or decision demands that are central to real-world driving. This gap calls for a framework that integrates events throughout the autonomy pipeline in a unified and scalable manner, rather than treating it as an isolated temporal cue.

To address this need, we introduce \eventdrive, a unified dataset and benchmark that integrates event streams, synchronized RGB frames, and language supervision across the \textbf{full driving loop}. Our goal is to move from passive sensing to actionable understanding within a coherent multimodal interface. We decompose driving into four sequential reasoning stages: \textbf{perception}, \textbf{understanding}, \textbf{prediction}, and \textbf{planning}, each expressed as a language-grounded task probing a complementary aspect of event-based reasoning. {Perception} assesses robustness under challenging illumination and motion, where events offer stable edges and temporal gradients that compensate for degraded RGB signals. {Understanding} targets object semantics and spatial relations, with asynchronous event cues helping to disambiguate interactions. {Prediction} evaluates short-term behavior anticipation, where the high temporal density of events exposes velocity and acceleration. {Planning} examines ego intent and waypoint estimation, leveraging continuous temporal structure for steadier decisions in dynamic environments. Together, these tasks form a unified evaluation protocol that highlights how temporal cues enhance perception and reasoning across the driving stack while remaining compatible with classical vision models and modern vision–language architectures.

Building upon \eventdrive, we develop \eventdrive-\textbf{VLM}, an event-driven vision–language model that integrates asynchronous event cues into unified multimodal reasoning. The framework tackles two core challenges in event-based driving. First, event streams vary widely in temporal density. To capture motion across scales, we introduce a \textbf{Dynamic Horizon Encoding module} that voxelizes events at multiple temporal resolutions and selects the most informative representation via a \textit{mixture-of-experts} gate. This preserves high-frequency dynamics during fast motion while ensuring stable aggregation in low-motion regimes. Second, event features must be aligned with the LLM’s semantic space. We address this with an \textbf{Event Q-Former Alignment module} that performs cross-attention between learnable event queries and temporally encoded representations, enabling motion-focused signals to integrate cleanly with language reasoning. A lightweight two-stage curriculum first learns event–language alignment with the LLM frozen, then performs instruction tuning to produce a coherent event-driven perception-to-action pipeline.

Comprehensive experiments on \eventdrive reveal strong modality contrasts. Frame-only VLMs perform well in perception but degrade sharply under low light and motion blur, and show limited spatial reasoning and weak motion inference. Event-only models excel in speed and direction prediction, highlighting the value of high-frequency temporal cues, yet lack semantic richness for appearance-heavy understanding. Our event–frame fusion model improves performance across all task families, stabilizing perception under adverse conditions, enhancing grounding and relational reasoning, and delivering stronger motion prediction and ego intent estimation. These results show that events and frames provide complementary strengths, and their integration yields more reliable multimodal reasoning than either alone.

In summary, our main contributions are as follows:

\begin{itemize}
\item We introduce \eventdrive, the first full-stack event and language benchmark for autonomous driving that unifies perception, understanding, prediction, and planning within a consistent multimodal framework.

\item We propose \eventdrive-VLM, a general training framework that equips large vision and language models with the ability to interpret, align, and reason over asynchronous event representations.

\item We establish a comprehensive evaluation protocol and extensively analyze how temporal cues improve multimodal reasoning, offering a foundation for future event-driven intelligence in real-world driving.

\end{itemize}

\section{Related Work}
\label{sec:relatedwork}

\begin{figure*}[t]
    \centering
    \includegraphics[width=0.95\linewidth]{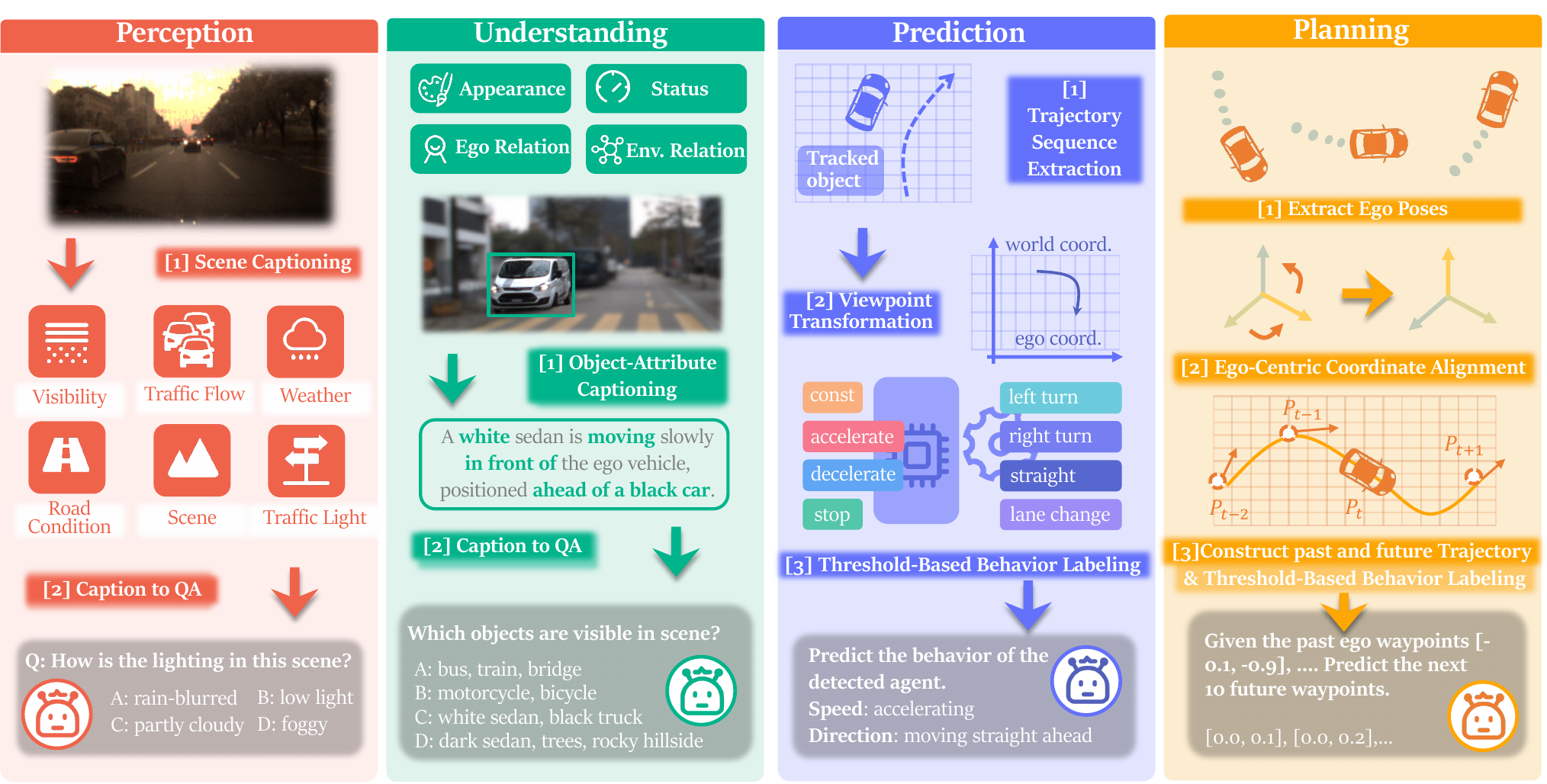}
    \caption{Annotation pipelines of \eventdrive. \includegraphics[width=0.016\linewidth]{icons/perception.png}~\textbf{Perception} converts scene-level attributes into structured QA; \includegraphics[width=0.016\linewidth]{icons/understanding.png}~\textbf{Understanding} generates object-level semantic captions and transforms them into QA; \includegraphics[width=0.016\linewidth]{icons/prediction.png}~\textbf{Prediction} extracts trajectories, applies ego-frame transformation, and assigns motion labels; \includegraphics[width=0.018\linewidth]{icons/planning.png}~\textbf{Planning} constructs ego-centric waypoints and produces corresponding decision-oriented supervision.
    }
    \label{fig:dataset_curation}
\end{figure*}

\noindent\textbf{Event Cameras for Driving.} 
Event sensors offer microsecond latency, high dynamic range, and robustness to motion blur, making them ideal for driving perception. They have been applied to detection \cite{gehrig2023rvt, lu2024flexevent, li2023sodformer, peng2024sast, wu2024leod}, segmentation \cite{kong2024openess, sun2022ess, jing2024hpl-ess}, tracking \cite{zhang2022spiking}, and recognition \cite{gehrig2019representation, kim2021n-imagenet, cho2023label}. Detection methods include graph- or spike-centric models preserving sparsity and asynchrony \cite{gehrig2022pushing, sun2023event, messikommer2020event, schaefer2022aegnn, cordone2022object, zhang2022spiking, cuadrado2023optical}, and dense feed-forward models that convert events into voxel grids or time surfaces for image-based backbones \cite{chen2018pseudo, iacono2018towards, jiang2019mixed, perot2020learning, li2022asynchronous, gehrig2023rvt, peng2024sast}. Beyond detection, event-based segmentation and tracking emphasize temporal continuity and motion cues \cite{sun2022ess, gehrig2020eklt, kong2025talk2event}. However, most systems remain limited to low-level perception without connecting to higher-level reasoning and planning. Our work advances this direction by integrating event sensing into a unified framework that links perception, understanding, and decision-making for end-to-end driving intelligence.

\noindent\textbf{Event-Frame Multimodal Learning.}
Events and RGB frames are complementary: events capture fine temporal dynamics under extreme lighting, while frames provide rich semantic context. Early pipelines relied on late fusion \cite{li2019event, chen2019multi}, whereas later designs introduced shared backbones and attention-based gating for cross-modal interaction \cite{tomy2022fusing, cao2022neurograsp, lu2024flexevent}. Transformer-based approaches further improved temporal reasoning and adaptive weighting under complex motion and illumination \cite{zhou2023rgb, li2023sodformer, wu2023eventclip}, benefiting tasks such as deblurring, depth estimation, and segmentation \cite{sun2022event, gehrig2021combining, hamaguchi2023hmnet}. Yet most frameworks rely on fixed temporal windows, which fail to capture motion cues that unfold over varying temporal scales \cite{li2023sodformer, cao2024embracing, gehrig2024dagr}. Our method addresses this by aligning multi-horizon event streams with frame semantics, enabling frequency-adaptive fusion that maintains both temporal precision and semantic coherence.

\noindent\textbf{Event-based VLMs.} 
While vision-language models (VLMs) have transformed visual understanding, extending them to asynchronous event data remains challenging \cite{kong2024openess,kong2025talk2event}. The key difficulty lies in embedding sparse, high-temporal-resolution signals into a shared vision-language space. Early contrastive approaches adapt CLIP-style alignment for zero-shot recognition \cite{wu2023eventclip, kong2024openess}, but rely on limited-scale datasets and rasterized inputs, reducing temporal fidelity. Recent works narrow this gap: EventGPT \cite{liu2024eventgpt} couples an event encoder and temporal aggregator with an LLM; EventVL \cite{li2025eventvl} trains on large-scale event-image-text pairs with spatiotemporal and semantic alignment; and LLaFEA \cite{zhou2025llafea} integrates frame-event fusion with cross-attention and duration cues. Despite these advances, most models remain confined to captioning or short QA tasks. On the contrary, our method preserves temporal sparsity and extends event-based VLMs toward comprehensive driving reasoning across diverse tasks.
\section{\eventdrive: A Vision–Language Benchmark for Event-Based Driving}
\label{sec:dataset}

In this section, we present the \eventdrive dataset and benchmark in detail. \cref{sec:dataset_overview} introduces the overall motivation and task hierarchy that define the dataset design. \cref{sec:data_curation} describes the annotation pipeline and data composition, highlighting how multimodal inputs are processed into language-grounded supervision. \cref{sec:metrics} summarizes the data splits and benchmark statistics, outlining the scale and coverage of the dataset and its comparison with existing event-based vision–language benchmarks.

\begin{table}[t]
\centering
\caption{
Statistics across four driving reasoning tasks. 
}
\resizebox{\linewidth}{!}{
\begin{tabular}{llrrr}
\toprule
\textbf{Task} & \textbf{Source} & \textbf{\#Train} & \textbf{\#Test} & \textbf{\#Hard} \\
\midrule
\multirow{3}{*}{\perception} 
 & DSEC \cite{gehrig2021dsec} & $53{,}196$ & $12{,}828$ & $780$ \\
 & M3ED \cite{chaney2023m3ed} & $65{,}280$ & $32{,}136$ & $15{,}396$ \\
 & PKU \cite{li2023sodformer} & $46{,}860$ & $11{,}400$ & $3{,}960$ \\
\midrule
\understanding & DSEC \cite{gehrig2021dsec} & $91{,}716$ & $28{,}848$ & $1{,}812$ \\
\prediction    & M3ED \cite{chaney2023m3ed} & $7{,}784$  & $2{,}810$  & $1{,}970$ \\
\planning      & M3ED \cite{chaney2023m3ed} & $46{,}290$ & $29{,}526$ & $18{,}951$ \\
\rowcolor{gray!10}
\multicolumn{2}{l}{\textbf{Total}} & $\mathbf{311{,}126}$ & $\mathbf{117{,}548}$ & $\mathbf{42{,}869}$ \\
\midrule
\rowcolor{gray!20}
\multicolumn{2}{l}{\textbf{Grand Total Samples}} & \multicolumn{3}{c}{$\mathbf{471,543}$} \\
\bottomrule
\end{tabular}
}
\label{tab:dataset}
\end{table}

\subsection{Hierarchical Task Framework}
\label{sec:dataset_overview}

Event cameras remain reliable under low light and rapid motion, offering advantages for safety-critical driving where conventional frames often fail. Existing event-based works focus mainly on isolated perception tasks, while RGB-based methods have progressed toward unified vision–language–action frameworks. To close this gap, we build a dataset that extends event understanding across the entire autonomy stack. The driving loop is organized into four sequential steps: environmental \textbf{perception}, object-centric \textbf{understanding}, forward \textbf{prediction}, and ego-driven \textbf{planning}. Each step is formulated as a language-grounded task that aligns event representations with multimodal models, enabling a systematic evaluation of how event signals contribute to high-level reasoning.

\noindent \perception describes global scene conditions across six subtasks: $^1$\textit{scene type}, $^2$\textit{visibility}, $^3$\textit{traffic flow}, $^4$\textit{weather}, $^5$\textit{traffic light}, and $^6$\textit{road condition}. Captions and structured QA pairs assess whether models can interpret environmental context from short temporal windows. Event streams preserve edge contrast and temporal gradients under challenging illumination or motion blur, strengthening robustness beyond frame-based perception.

\noindent \understanding captures object-level semantics and spatial structure using captions, QA pairs, and bounding boxes spanning $^1$\textit{object presence}, $^2$\textit{appearance}, $^3$\textit{motion state}, $^4$\textit{ego relation}, $^5$\textit{environment relation}, and $^6$\textit{grounding}. The asynchronous event signal supplies fine temporal cues that support reasoning about motion and spatial relations that static frames often miss.

\noindent \prediction targets short-horizon behavioral forecasting through indicators of $^1$\textit{speed} change and $^2$\textit{direction} change. Each instance uses language queries to evaluate whether models can infer future motion trends of surrounding objects from limited observations. The high temporal density of events reveals velocity and acceleration directly, improving the fidelity of motion forecasting.

\noindent \planning evaluates ego-centric decision making using $^1$\textit{speed intent}, $^2$\textit{direction intent}, and $^3$\textit{waypoint planning}. Models must convert multimodal observations into coherent driving decisions. Event sensing preserves continuous awareness of dynamic changes, supporting more stable reasoning in rapidly evolving or low-visibility conditions.

\subsection{Language-Grounded Data Generation}
\label{sec:data_curation}
We construct \eventdrive through a semi-automatic pipeline built upon the event datasets DSEC~\cite{gehrig2021dsec}, M3ED~\cite{chaney2023m3ed}, and PKU-DAVIS-SOD~\cite{li2023sodformer}, which span diverse environments and illumination conditions. Using synchronized RGB frames, event streams, bounding boxes, LiDAR, and ego-pose signals, we employ Qwen3-VL~\cite{qwen3} to generate linguistically structured supervision across the four task dimensions. The pipeline, shown in \cref{fig:dataset_curation}, integrates automatic captioning with controlled question–answer generation to ensure both scalability and semantic consistency.

Scene-level \textbf{perception} labels are produced by prompting Qwen3-VL to generate global captions describing the six environmental attributes, which are subsequently decomposed into balanced question–answer pairs with explanatory sentences. Object-level \textbf{understanding} builds on ground-truth bounding boxes from DSEC. Qwen3-VL first generates descriptions for the five object attributes, and these descriptions are transformed into visual question–answer and grounding tasks that link textual queries to specific spatial regions. For \textbf{prediction}, we use ego-pose to project 3D boxes from~\cite{liang2025perspective} into the ego frame, extract object trajectories, and convert their motion patterns into natural-language descriptions of speed and direction intent, paired with corresponding QA items. Ego-level \textbf{planning} leverages M3ED trajectory supervision to derive speed intent, path intent, and future waypoints, which are aligned with language queries that evaluate ego decision making. Together, these components yield a unified annotation framework that connects perception, understanding, prediction, and planning through temporally and semantically coherent language-grounded supervision.

\begin{figure*}[t]
    \centering
    \includegraphics[width=0.95\linewidth]{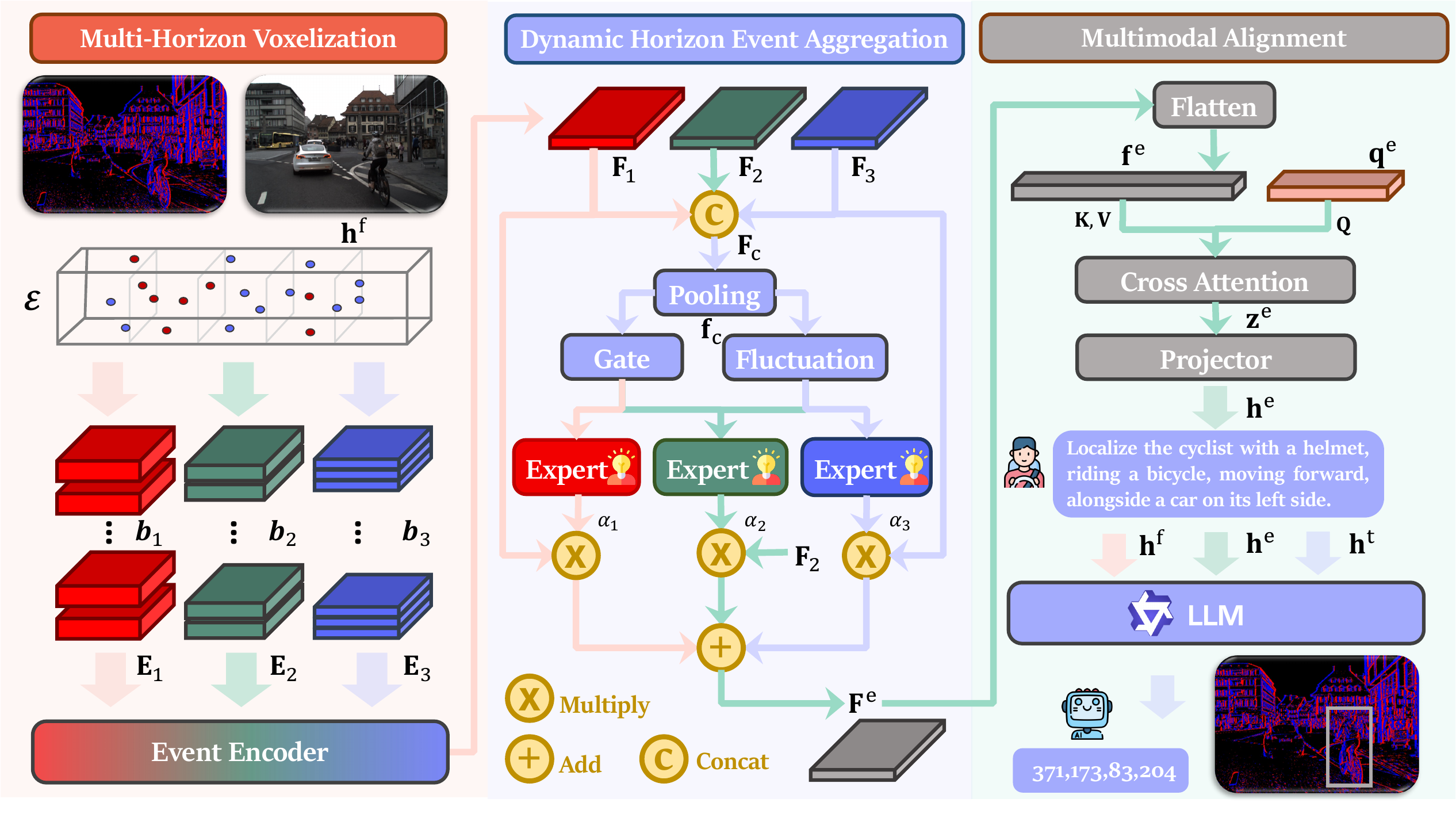}
    \caption{
    \textbf{\eventdrive-VLM Overview.} We first convert asynchronous events into \textbf{multi-horizon voxel tensors} that capture motion at \textit{different temporal scales}. A \textbf{dynamic horizon event encoder} then aggregates these representations through a \textit{Mixture-of-Experts} gating mechanism (\cf~\cref{sec:event_encoder}). An \textbf{Event Q-Former} performs cross-attention to extract language-aligned, motion-aware tokens, enabling coherent fusion of multi-modal features within the LLM for unified driving reasoning (\cf~\cref{sec:qformer}). A \textbf{two-stage training curriculum} further aligns event, visual, and linguistic pathways, strengthening cross-modal grounding and downstream reasoning performance (\cref{sec:training}).
    }
    \label{fig:framework}
\end{figure*}

\subsection{Compositional Structure and Statistics}
\label{sec:metrics}

We adopt standard \textit{training} and \textit{testing} splits and further introduce a \textbf{hard} split consisting solely of \textit{low-light} and \textit{motion-blur} sequences. This split enables targeted evaluation of the advantages of event sensing under conditions where frame-based perception degrades. As shown in \cref{tab:dataset}, the dataset contains $\mathbf{471{,}543}$ event–frame–text samples, offering large-scale multimodal supervision across perception, understanding, prediction, and planning. Compared with existing event–language datasets~\cite{liu2024eventgpt, zhou2025llafea, li2025eventvl}, which often rely on simulated data or provide limited real-world coverage below $100$k samples, \eventdrive offers a significantly broader and more diverse set of real-world annotations. It supports a wider range of tasks that connect event-driven perception with higher-level reasoning and decision-making, forming a more comprehensive benchmark for evaluating multimodal driving intelligence.
\section{\eventdrive-VLM: A Unified Model for Event-Based Driving Intelligence}
\label{sec:method}

Building upon \eventdrive, our goal is to develop an event-driven vision–language–action model that can interpret and generate decisions grounded in the spatio-temporal structure of event streams. As shown in \cref{fig:framework}, the core of \eventdrive-VLM is a multimodal LLM backbone that integrates asynchronous event cues with pretrained visual–linguistic knowledge to support coherent reasoning across driving tasks. To bridge event dynamics with the LLM embedding space, the model incorporates three components: a \textbf{dynamic horizon event encoder} that adapts to varying sampling frequencies and motion patterns (\cf~\cref{sec:event_encoder}), an \textbf{Event Q-Former} that extracts language-aligned and motion-aware representations (\cf~\cref{sec:qformer}), and a \textbf{two-stage training curriculum} that progressively aligns event, visual, and linguistic pathways for unified multimodal reasoning (\cf~\cref{sec:training}).

\subsection{Dynamic Horizon Event Encoder}
\label{sec:event_encoder}

Event data exhibit large variation in temporal density across datasets and tasks. Sensors operate at different sampling rates, and the required temporal resolution also depends on the objective: perception benefits from broader temporal context, whereas prediction and planning rely on fine-grained motion cues. Conventional voxelization~\cite{gehrig2023rvt} applies a fixed number of temporal bins and thus compresses long exposure windows while blurring fast motion, losing the high-frequency detail essential for motion reasoning.

To address this issue, we propose a \textbf{dynamic horizon encoding} strategy that adapts to varying time scales and scene dynamics. Given an event stream $\mathcal{E} = \{e_k\}_{k=1}^{K}$, where each event $e_k = (x_k, y_k, t_k, p_k)$ encodes spatial coordinates, timestamp, and polarity $p_k \in \{-1, 1\}$, standard voxelization maps events into a 4D tensor $\mathbf{E} \in \mathbb{R}^{2 \times B \times H \times W}$:
\begin{equation}
\mathbf{E}(p, \tau, x, y) = \sum_{e_k\in\mathcal{E}}\delta(p - p_k)\delta(x-x_k,y-y_k)\delta(\tau-\tau_k),
\label{eq:events}
\end{equation}
where $\tau_k = \left\lfloor \frac{t_k - t_a}{t_b - t_a} B \right\rfloor$. Instead of relying on a single bin size $B$, we construct multiple voxel tensors using temporal resolutions $\mathcal{B} = \{ b_n \}_{n=1}^{N}$, producing event tensors $\mathbf{E}_n$ that capture short-, medium-, and long-horizon motion patterns.

To adaptively select the most suitable temporal horizon for different motion patterns and task requirements, we employ a \textit{Mixture-of-Experts (MoE)}~\cite{shazeer2017outrageously} gating mechanism that dynamically weights multi-scale temporal experts according to scene dynamics. Each expert network~\cite{gehrig2023rvt} specializes in one temporal resolution, processing a voxelized tensor $\mathbf{E}_n$ with bin size $b_n$ to produce an encoded feature
\begin{equation}
    \mathbf{F}_n = \boldsymbol{\sigma}(\mathbf{E}_n),
    \qquad 
    \mathbf{F}_n \in \mathbb{R}^{H \times W \times d},
\end{equation}
where $d$ is the channel dimension. The resulting expert features are concatenated into a contextual representation $\mathbf{F}_c \in \mathbb{R}^{H \times W \times (Nd)}$. Global average pooling produces a compact descriptor $\mathbf{f}_c \in \mathbb{R}^{Nd}$ that summarizes multiscale temporal cues. The gating logits are then computed as
\begin{equation}
    \mathbf{z}
    = \mathbf{W}_g \mathbf{f}_c
    + \texttt{Softplus}\!\big(
        \boldsymbol{\epsilon}
        \odot (\mathbf{W}_{\text{noise}} \mathbf{f}_c)
      \big),
\end{equation}
where 
$\mathbf{W}_g, \mathbf{W}_{\text{noise}} \in \mathbb{R}^{Nd \times N}$ 
are trainable parameters, and 
$\boldsymbol{\epsilon} \sim \mathcal{N}(0, 1)$ 
introduces controlled stochasticity to encourage expert diversity. Retaining only the largest $k$ logits focuses computation on the most relevant temporal experts, and softmax normalization yields the weights $\alpha_n$. The aggregated event representation is computed as $\mathbf{F}^e = \sum_{n=1}^{N} \alpha_n \, \mathbf{F}_n$.
This adaptive formulation allows the encoder to emphasize high-resolution temporal features when motion is fast and to leverage coarse but stable aggregation when dynamics are mild. As a result, the model maintains temporal fidelity across a wide range of driving scenarios while remaining computationally efficient.

\subsection{Event Q-Former Alignment}
\label{sec:qformer}
After obtaining the aggregated event representation $\mathbf{F}^e$, we align it with the LLM embedding space using an \textbf{Event Q-Former (EQA)}. Rather than concatenating event and frame tokens, which ignores modality asymmetry and leads to high computational cost, the EQA employs cross-attention to extract the motion-relevant components of the event representation that matter for language-guided reasoning.

Following the Q-Former architecture~\cite{li2023blip}, we introduce a set of learnable event query tokens $\mathbf{q}^e \in \mathbb{R}^{N_q \times d}$, which attend to the event feature map $\mathbf{F}^e \in \mathbb{R}^{H \times W \times d}$. After flattening $\mathbf{F}^e$ into $\mathbf{f}^e \in \mathbb{R}^{(HW) \times d}$, the attended event embeddings are computed as
\begin{equation}
    \mathbf{z}^e = 
    \texttt{softmax}\!\left(
        \frac{ (\mathbf{q}^e \mathbf{W}_Q) ( \mathbf{f}^e \mathbf{W}_K )^\top }
             { \sqrt{d} }
    \right)
    ( \mathbf{f}^e \mathbf{W}_V ),
\end{equation}
where $\mathbf{W}_Q, \mathbf{W}_K, \mathbf{W}_V \in \mathbb{R}^{d \times d}$ are learnable projections. The output $\mathbf{z}^e \in \mathbb{R}^{N_q \times d}$ forms a compact representation that summarizes the most salient temporal and motion cues.

The Event Q-Former enables each query to attend selectively to temporally informative regions in the event stream, producing motion-aware embeddings suitable for downstream multimodal reasoning. This structured attention preserves the temporal distinctiveness of events while supporting coherent interaction with frame and text representations. A lightweight projection layer maps the attended event features into the LLM embedding space, yielding event tokens $\mathbf{h}^e$ that align with the projected frame tokens $\mathbf{h}^f$ and text embeddings $\mathbf{h}^t$. The concatenated sequence is then used as the input to the LLM decoder, enabling unified multimodal reasoning across all driving tasks.

\begin{table*}[t]
\centering
\caption{
\textbf{Comparison across four driving–reasoning tasks} in the \eventdrive benchmark. For \textbf{perception}, we report \texttt{QA Accuracy} on three subsets. For \textbf{understanding}, we report \texttt{QA Accuracy}, grounding Top-1 \texttt{Accuracy} at IoU $0.6$, and \textbf{mIoU}. For \textbf{prediction} and \textbf{planning}, we report \texttt{Speed Accuracy} and \texttt{Path Accuracy}, and for planning, we also report the mean \textbf{L2 Error} (lower is better). All scores are reported in percentage (\%), except for L2 Error in meters. Best results are shown in \textbf{bold} and 2nd-best results are \underline{underlined}.
}
\resizebox{\linewidth}{!}{
\begin{tabular}{r|ccc|ccc|cc|ccc}
\toprule
\multirow{2}{*}{\textbf{Method}} & 
\multicolumn{3}{c|}{\includegraphics[width=0.022\linewidth]{icons/perception.png}~\textbf{Perception}} & 
\multicolumn{3}{c|}{\includegraphics[width=0.022\linewidth]{icons/understanding.png}~\textbf{Understanding}}  & 
\multicolumn{2}{c|}{\includegraphics[width=0.022\linewidth]{icons/prediction.png}~\textbf{Prediction}}  & 
\multicolumn{3}{c}{\includegraphics[width=0.022\linewidth]{icons/planning.png}~\textbf{Planning}}  \\
 & \texttt{Acc@D} & \texttt{Acc@M} & \texttt{Acc@P}
 & \texttt{Acc} & \texttt{Acc@60} & \textbf{mIoU} 
 & \texttt{Speed} & \texttt{Path} 
 & \texttt{Speed} & \texttt{Path} & \textbf{L2 Error} \\
\midrule\midrule
\rowcolor{event_green!15}\multicolumn{12}{c}{\textcolor{event_green}{\textbf{Event-based Models}}} 
\\
EventGPT-7B \cite{liu2024eventgpt} & $51.40$  & $54.97$ & $52.25$  & $38.78$ & $5.49$  & $8.24$  & $27.84$  &  $65.44$ & $28.99$  & $76.08$  & $11.42$  \\
\eventdrive-VLM      & $62.51$  & $67.07$  & $62.39$  &$54.21$ & $43.43$  & $47.82$   & $34.68$  & $82.25$  &  $42.56$ & $84.64$  & $6.89$  \\
\midrule
\rowcolor{frame_red!15}\multicolumn{12}{c}{\textcolor{frame_red}{\textbf{Frame-based Models}}} 
\\
LLaVA-v1.6-Mistral-7B-hf \cite{llavamistral} & $58.65$ & $59.24$ & $50.05$  & $40.37$  & $15.02$ & $29.0$  & $18.22$  & $29.68$  &  $9.89$ & $87.51$  & $8.13$  
\\
LLaVA-OneVision-1.5-8B \cite{llavaonevision} & $80.81$ & \underline{$86.35$} & $72.42$  & \underline{$61.62$} & $2.91$  & $12.5$  &  $14.38$ & $83.91$  & \underline{$56.51$}  & $58.50$  & $9.19$  
\\
InternVL2.5-8B \cite{chen2024expanding} & $78.42$ & $83.41$ & $75.86$  & $60.02$ & $0.29$  & $2.5$  &  \underline{$39.72$} & $86.41$  & $12.52$  & $73.34$  & $10.19$  
\\
InternVL3-8B \cite{zhu2025internvl3} & $78.61$ & $86.27$ & $74.37$ & $60.60$ &  $0.24$ & $2.31$  &  $4.41$ & $\mathbf{91.25}$  & $52.04$  &  $84.34$ &  $9.84$ \\
Qwen2.5-VL-7B-Instruct \cite{qwen2.5-VL} & $77.71$ & $76.92$ & $62.46$ & $49.98$ & $63.35$ & $60.11$ & $8.47$ & $81.64$ & $39.32$ & $87.41$ & $8.78$ 
\\
Qwen2.5-VL-7B-Instruct* \cite{qwen2.5-VL} & \underline{$81.52$}  & $84.69$  & \underline{$75.88$}  & $58.44$  & \underline{$69.94$}  & \underline{$67.19$}  & $36.84$  &  $84.94$ & $53.87$  & \underline{$89.44$} & \underline{$4.54$} 
\\
\midrule
\rowcolor{fusion_blue!15}\multicolumn{12}{c}{\textcolor{fusion_blue}{\textbf{Event + Frame Models}}} 
\\
\eventdrive-VLM   & $\mathbf{85.44}$ & $\mathbf{86.64}$ & $\mathbf{78.89}$  & $\mathbf{65.46}$  & $\mathbf{72.86}$  & $\mathbf{72.56}$  & $\mathbf{42.44}$   &  \underline{$87.49$}  & $\mathbf{57.03}$   &  $\mathbf{92.35}$  & $\mathbf{3.66}$   \\
\bottomrule
\end{tabular}
}
\label{tab:eventdrive}
\end{table*}

\subsection{Training Curriculum}
\label{sec:training}
We adopt a two-stage training curriculum that progresses from event–language grounding to multimodal instruction following, enabling stable and efficient adaptation of the event-driven VLM.

\noindent\textbf{Event–Language Pre-adaptation.}
In the first stage, the LLM and the frame visual encoder are kept frozen, while the event encoder, Q-Former, and projection layers are trained with a language modeling objective on paired caption data. Although frame data are available, gradients flow only through the event pathway, allowing the model to form a linguistically aligned event representation without altering the pretrained frame semantics. This generative supervision encourages the event encoder and Q-Former to organize temporal cues and motion structure into embeddings compatible with the LLM space, providing a stable form of cross-modal alignment.

\noindent\textbf{Instruction Tuning.}
In the second stage, we unfreeze the transformer blocks of the LLM and fine-tune the entire event pathway together with these LLM layers on all caption and QA data, while keeping the frame visual encoder frozen. This phase integrates temporal and semantic signals more tightly, enabling consistent reasoning across perception, understanding, prediction, and planning. Joint optimization under multimodal instructions yields coherent grounding between event dynamics and textual responses, forming a unified event-driven perception-to-action model.

\section{Experiment}
\label{sec:experiment}

\subsection{Experimental Settings}

\noindent\textbf{Implementation Details.}
We fine-tune Qwen2.5-VL-7B-Instruct~\cite{qwen2.5-VL} with a pretrained RVT~\cite{gehrig2023rvt} backbone as the event encoder. Dynamic horizon encoding uses temporal bins $\mathcal{B}={20,50,100}$, and the event projector is a linear layer. The RGB visual tower remains frozen. Training adopts AdamW~\cite{kingma2015adam} with a cosine schedule on 16 NVIDIA H20 GPUs, using a learning rate of $1\times10^{-4}$ for the event encoder and projector. Mixed precision bf16 is used throughout. Pre-adaptation and instruction tuning each run for two epochs with a batch size of $128$. Packed sequences and flattened multimodal inputs are applied with a sequence length of $4096$ tokens, and FlashAttention 2~\cite{dao2023flashattention} accelerates all attention layers. 

\noindent\textbf{Evaluation Metrics.}
We establish a unified evaluation protocol across all defined tasks. Perception is evaluated by QA accuracy over six scene attributes. Understanding combines QA accuracy across five object-level aspects with grounding metrics, including Top-1 localization accuracy at IoU $0.6$ and mean IoU. Prediction assesses speed and path intent accuracy. Planning evaluates intent accuracy together with mean L2 waypoint error across $1$, $3$, and $5$ seconds. Additional details appear in the supplementary materials.

\noindent\textbf{Baselines.}
Comparisons include both \textit{frame-only} and \textit{event-only} models. Frame-only evaluation covers Qwen2.5 VL \cite{qwen2.5-VL}, InternVL \cite{zhu2025internvl3, chen2024expanding}, and LLaVA \cite{llavaonevision, llavamistral} under zero-shot inference, which tests robustness in low light and motion-blurred scenes. To isolate the contribution of events, we additionally fine-tune a Qwen2.5-VL-7B model using the same instruction tuning protocol as our model. For event-only comparison, we evaluate the open-sourced EventGPT in a zero-shot manner since its training code is unavailable. We further report zero-shot results on the released portion of the Event-Chat~\cite{liu2024eventgpt} dataset, noting that the absence of an official split limits this evaluation to reference use.

\begin{table}[t]
\centering
\caption{
\textbf{Comparative results on the Event-Chat dataset~\cite{liu2024eventgpt}}, which evaluates event-driven description/reasoning through \textbf{Detailed Captioning}, \textbf{Complex Reasoning}, and \textbf{Visual Question Answering}. Higher values indicate better performance.
}
\label{tab:eventgpt}
\resizebox{\linewidth}{!}{ 
\begin{tabular}{r|c|cccc}
\toprule
{\textbf{Models}} & \textbf{Params} & \textbf{DC} & \textbf{CR} & \textbf{VQA} \\
\midrule\midrule
\rowcolor{frame_red!15}\multicolumn{5}{c}{\textcolor{frame_red}{\textbf{Frame-based Models}}} 
\\
LLaVA-7B-v1.5 \cite{liu2023llava}       & 7B & $2.20$ & $4.04$ & $3.26$ \\
Qwen2-VL-7B \cite{Qwen2VL}              & 7B & $2.38$ & $4.02$ & $2.91$ \\
InternVL2-8B \cite{chen2024internvl}   & 8B & $2.37$ & $4.00$ & $3.71$ \\
Deepseek-vl-7b \cite{lu2024deepseekvl}  & 7B & $2.41$ & \underline{$4.10$} & $3.37$ \\
\rowcolor{event_green!15}\multicolumn{5}{c}{\textcolor{event_green}{\textbf{Event-based Models}}} 
\\
EventGPT-7B \cite{liu2024eventgpt}      & 7B & $\mathbf{3.52}$ & $4.09$ & $\mathbf{4.29}$\\
\eventdrive-VLM                & 7B & \underline{$3.43$} &  $\mathbf{4.15}$ & \underline{$3.94$}   \\
\bottomrule
\end{tabular}
}
\end{table}

\subsection{Comparative Studies}

\noindent\textbf{Comparisons with Event-Only Methods.}
Event-only models in \cref{tab:eventdrive} perform well on motion-centric tasks such as speed and path intent, confirming that high-frequency temporal cues are inherently well suited for short-horizon motion reasoning. They also show reasonable grounding performance despite lacking explicit spatial supervision, highlighting the structural information encoded in event streams. Yet, the absence of appearance cues limits their semantic understanding, resulting in lower perception and object-centric accuracy. Results on the Event-Chat benchmark (\cref{tab:eventgpt}) further show that the representations learned from \eventdrive transfer across datasets, indicating that the dataset promotes generalizable event–language alignment rather than overfitting to a specific annotation style.

\noindent\textbf{Comparisons with Frame-Only Methods.}
Frame-based VLMs achieve high accuracy on perception under normal conditions but degrade sharply in \texttt{Acc@P}, understanding \texttt{Acc@60}, and speed-related prediction metrics due to their inability to encode motion or handle low-light and blur. These limitations lead to unstable spatial reasoning and poor motion intent inference. Introducing events alleviates these issues: event–frame fusion improves all metrics, especially grounding and speed intent, showing that temporal gradients and motion cues are essential for reliable reasoning and cannot be inferred from RGB imagery alone.

\begin{figure}[t]
    \centering
    \includegraphics[width=0.97\linewidth]{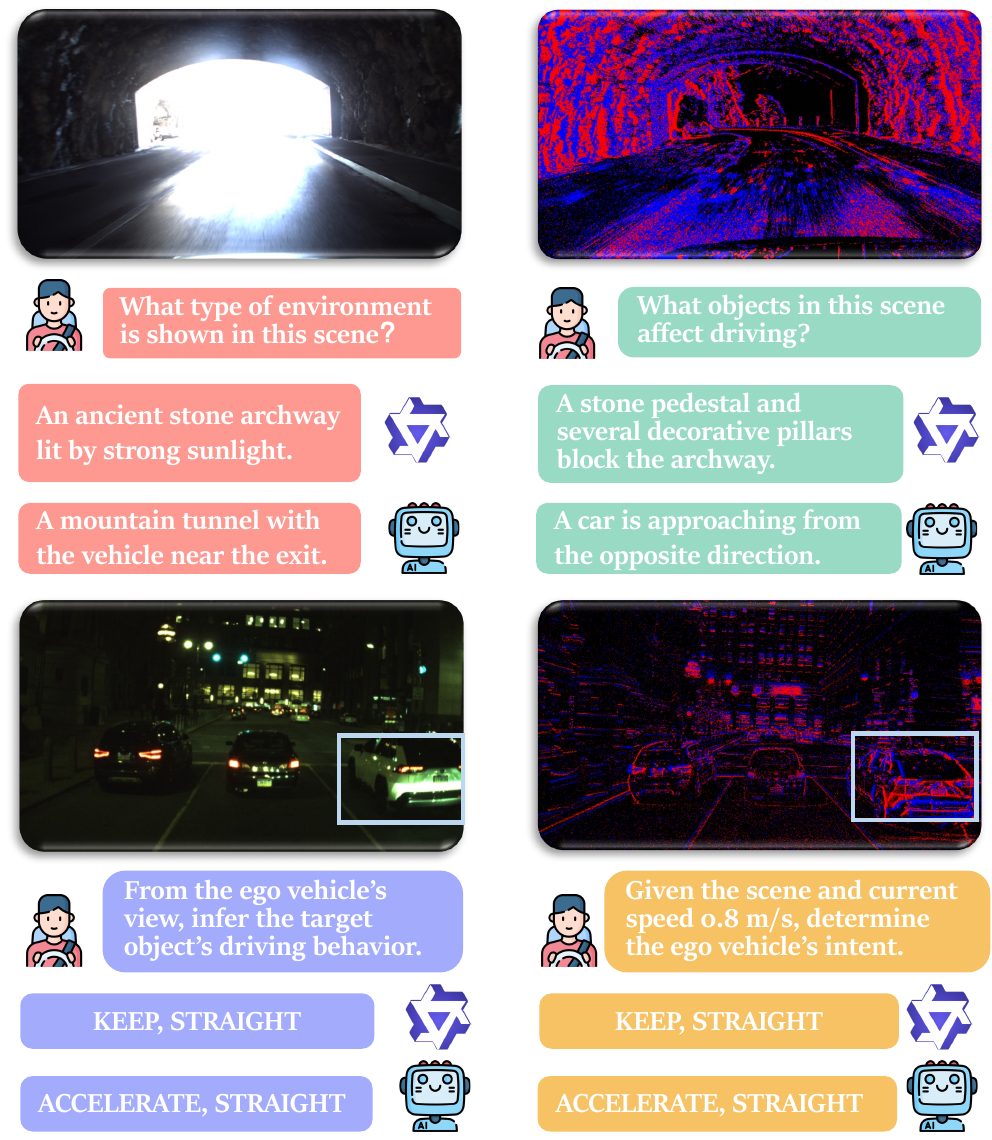}
    \caption{Qualitative results on \eventdrive comparing \eventdrive-VLM with Qwen. Events remain reliable under low light and motion, improving scene perception, object understanding, motion prediction, and ego intent estimation.
    }
    \label{fig:result}
\end{figure}

\noindent\textbf{Comparisons across Different Tasks.}
The four task families reveal complementary strengths of different modalities. Perception is broadly solvable, but event streams offer improved robustness under visual degradations. Understanding exposes significant variance across VLMs, as many struggle with spatial relations and grounding from a single RGB frame; temporal cues help resolve these ambiguities. Prediction presents the largest modality gap: inferring speed or direction from static frames is ill-posed, whereas events directly encode motion, leading to consistently higher accuracy for event-enhanced models. Planning mirrors this pattern, where temporal dynamics from traffic improve speed intent estimation and reduce trajectory L2 error. These trends collectively demonstrate that events provide structural and temporal information that complements semantics from frames across the full driving chain. A qualitative comparison is shown in \cref{fig:result}.

\subsection{Ablation Studies}

\begin{table}[t]
\centering
\caption{Ablation results on the \eventdrive benchmark. We report average \texttt{QA Accuracy} for perception, \texttt{QA Accuracy} and \textbf{mIoU} for understanding, and mean \texttt{Speed}/\texttt{Path Accuracy} for prediction and planning. Planning additionally includes mean \textbf{L2 Error} in meters (lower is better).
}
\resizebox{\linewidth}{!}{
\begin{tabular}{r|c|cc|c|cc}
\toprule
\multirow{2}{*}{\textbf{Method}} & 
\multicolumn{1}{c|}{\includegraphics[width=0.044\linewidth]{icons/perception.png}~\textbf{Per.}} & 
\multicolumn{2}{c|}{\includegraphics[width=0.044\linewidth]{icons/understanding.png}~\textbf{Und.}}  & 
\multicolumn{1}{c|}{\includegraphics[width=0.044\linewidth]{icons/prediction.png}~\textbf{Pre.}}  & 
\multicolumn{2}{c}{\includegraphics[width=0.044\linewidth]{icons/planning.png}~\textbf{Plan.}}  \\
 & \texttt{Acc}
 & \texttt{Acc} & \textbf{mIoU} 
 & \texttt{Acc}  
 & \texttt{Acc} & \textbf{L2 Error} \\
\midrule\midrule
\rowcolor{event_green!15}\multicolumn{7}{c}{\textcolor{event_green}{\textbf{Voxelization}}} 
\\
$\mathbf{N=1}$ & $82.40$  & $62.33$ & $69.52$  & $63.96$ & $71.18$  & $4.11$ 
\\
$\mathbf{N=5}$ & $\mathbf{83.95}$  & $64.97$ & \underline{$72.25$}  & $62.78$ & $73.49$  & $3.88$  
\\
\midrule
\rowcolor{frame_red!15}\multicolumn{7}{c}{\textcolor{frame_red}{\textbf{Dynamic Horizon Event Aggregation}}}
\\
\textbf{Add} & $76.76$  & $63.89$ & $67.64$  & $62.50$ & $72.49$  & $4.57$ 
\\
\textbf{Wt.sum} & \underline{$83.84$}  & $64.67$ & $70.56$  & $61.08$ & $74.10$  & $3.75$ 
\\
\midrule
\rowcolor{fusion_blue!15}\multicolumn{7}{c}{\textcolor{fusion_blue}{\textbf{Multimodal Alignments}}} 
\\
\textbf{Concat} & $79.35$  & $61.08$ & $71.93$  & \underline{$64.76$} & $72.62$  & $4.01$ 
\\
\textbf{Attention} & $81.25$  & \underline{$65.12$} & $70.23$  & $62.14$ & $\mathbf{75.85}$  & \underline{$3.69$} 
\\
\midrule
Ours   & $83.66$  & $\mathbf{65.46}$  & $\mathbf{72.56}$  & $\mathbf{64.96}$  & \underline{$74.69$}  & $\mathbf{3.66}$   \\
\bottomrule
\end{tabular}
}
\label{tab:ablations}
\end{table}

\noindent\textbf{Multi-Horizon Voxelization.}
We evaluate the impact of using different numbers of temporal resolutions when voxelizing event streams. As shown in \cref{tab:ablations}, increasing the number of horizons from one to five improves performance across all tasks, particularly in understanding and planning, where temporal diversity helps disambiguate object motion and spatial relations. However, the performance gain saturates beyond three horizons, while computational cost grows proportionally with the number of voxelized tensors. Our choice ($N=3$), therefore, provides the best trade-off between accuracy and efficiency, capturing both short-range motion cues and longer-term temporal context without incurring unnecessary overhead.

\noindent\textbf{Dynamic Horizon Event Aggregation.}
We compare different strategies for aggregating multi-horizon event features. A naïve summation (“Add”) collapses temporal distinctions and yields the weakest results, confirming that indiscriminate fusion fails to preserve horizon-specific information. Weighted summation (“Wt.sum”) improves performance by allowing the model to favor informative horizons, but still underperforms compared with selecting a single expert. Our MoE-based dynamic horizon module, which activates only the top-scoring expert, achieves the best overall results, indicating that event representations benefit from horizon specialization and that suppressing irrelevant resolutions is more effective than blending them.

\noindent\textbf{Multimodal Alignment.}
We study the alignment between event features and the multimodal embedding space. Simple concatenation of event and frame tokens leads to suboptimal performance due to modality imbalance and increased sequence length. Cross-attention improves grounding and planning by enabling token-level interactions, yet remains inferior to our Event Q-Former. The Q-Former achieves the best results while maintaining a lower computational cost, as learnable queries extract only the most salient motion patterns rather than attending to the full spatiotemporal map. This confirms that structured, query-centric alignment offers both efficiency and stronger motion abstraction.
\section{Conclusion}
\label{sec:conclusion}

In this work, we introduce \eventdrive and \eventdrive-VLM to unify event sensing, frames, and language across the full driving stack. Our results show that high-frequency temporal cues provide complementary strengths to RGB and significantly enhance multimodal perception, reasoning, and trajectory forecasting. Event–frame fusion improves robustness and motion understanding, highlighting events as a key modality for future driving systems.

\section*{Acknowledgments}

This work is under the programme DesCartes and is supported by the National Research Foundation, Prime Minister’s Office, Singapore, under its Campus for Research Excellence and Technological Enterprise (CREATE) programme. The authors also gratefully acknowledge Horizon Robotics for its support and computational resources.

The authors would like to sincerely thank the Program Chairs, Area Chairs, and Reviewers for the time and effort devoted during the review process.
\vspace{0.1cm}

\appendix
\renewcommand{\appendixname}{Appendix~\Alph{section}}

\section*{Appendix}
\startcontents[appendices]
\printcontents[appendices]{l}{1}{\setcounter{tocdepth}{3}}
\vspace{0.3cm}

\section{\eventdrive Dataset}

\eventdrive is a unified event-frame driving benchmark designed to support the full spectrum of driving reasoning, spanning \textbf{perception}, \textbf{understanding}, \textbf{prediction}, and \textbf{planning}. It integrates heterogeneous multimodal data from three complementary sources: M3ED, DSEC, and PKU-DAVIS-SOD, covering diverse illumination regimes, motion patterns, and annotation types. This section first introduces the source datasets and their sensing characteristics, then describes our cross-dataset construction strategy for forming coherent splits for each reasoning level. Finally, we outline the annotation pipeline that converts raw multimodal streams into structured, language-driven supervision.

\subsection{Source Datasets}

A unified event-frame driving benchmark requires diverse sensing conditions, high-speed motion patterns, and dense multimodal annotations. To construct \eventdrive, we draw from three complementary datasets: M3ED \citet{chaney2023m3ed}, DSEC \cite{gehrig2021dsec}, and PKU-DAVIS-SOD \cite{li2023sodformer}, each providing unique combinations of resolution, temporal fidelity, scene diversity, and supervision quality. A comparison of these datasets is shown in \cref{tab:source_datasets}. Together, they offer a wide range of illumination and motion regimes and rich annotations spanning 2D/3D perception, tracking, and segmentation, forming a comprehensive foundation for event-driven multimodal reasoning.

\subsubsection{{M3ED Dataset}}

The \textbf{M3ED dataset} is a large-scale multimodal benchmark designed for high-speed and high-dynamic robotic perception across diverse environments. Unlike prior event-camera datasets that target a single platform or homogeneous driving scenes, M3ED spans \textbf{multiple robotic platforms}, including an autonomous ground vehicle, a quadruped robot, and a UAV, all equipped with a unified sensor suite. Each platform streams synchronized Prophesee Gen4 event cameras ($1280 \times 720$), global-shutter grayscale and RGB cameras ($1280 \times 800$), a $64$-beam Ouster LiDAR, and a high-quality VectorNav IMU, enabling consistent multimodal perception under extreme motion and illumination changes.

\begin{table*}[t]
\centering
\caption{
Comparison of the three source datasets used to construct \eventdrive. 
All datasets provide synchronized RGB frames and event streams with diverse environments and motion characteristics.
}
\resizebox{\linewidth}{!}{
\begin{tabular}{lccccccc}
\toprule
\textbf{Dataset} & \textbf{\#Seq} & \textbf{\#Time (s)} & \textbf{Resolution} & \textbf{FPS} & \textbf{\#Frames} & \textbf{Annotations} & \textbf{Domain} \\
\midrule
\rowcolor{gray!05}
\textbf{M3ED} \cite{chaney2023m3ed} 
 & $70$ 
 & $12,237$
 & $1280\!\times\!720$ 
 & $30$ Hz 
 & $>\!300{,}000$ 
 & Pose, depth, 2D/3D seg. 
 & Driving, UAV, quadruped 
\\
\textbf{DSEC} \cite{gehrig2021dsec, gehrig2024dagr}
 & $60$ 
 & $\sim3,600$
 & $640\!\times\!480$ 
 & $20$ Hz 
 & $78{,}344$ 
 & Depth, flow, 2D boxes/seg.
 & Driving
\\
\rowcolor{gray!05}
\textbf{PKU-DAVIS-SOD} \cite{li2023sodformer}
 & $220$ 
 & $\sim11,000$
 & $346\!\times\!260$ 
 & $25$ Hz 
 & $\sim276{,}000$ 
 & 2D boxes 
 & Driving 
\\
\bottomrule
\end{tabular}
}
\label{tab:source_datasets}
\end{table*}

The dataset contains $\mathbf{70}$ sequences collected across urban streets, indoor corridors, forest trails, and aerial environments, totaling more than 3 TB of data. These sequences cover a wide range of dynamic conditions: high-speed car driving in dense traffic and tunnels, aggressive UAV flights through cluttered vegetation and urban canyons, and quadruped locomotion with gait-induced vibrations on uneven terrains. Such diversity produces extremely high event rates, often exceeding 200 MEPS during rapid rotations or high-texture scenes, making M3ED a challenging benchmark for temporal reasoning and motion-aware perception.

M3ED provides rich ground-truth supervision for multimodal learning. This includes \textbf{global pose trajectories} estimated by FasterLIO, \textbf{dense depth maps} computed from accumulated LiDAR sweeps, \textbf{2D semantic segmentation} with eleven categories aligned with the DSEC taxonomy, and \textbf{3D instance annotations} for pedestrians, vehicles, buildings, and trees. The dataset additionally supports evaluation of event-based optical flow and odometry, facilitating research across both low-level motion estimation and high-level semantic understanding.

By combining heterogeneous platforms, synchronized multimodal sensors, and geographically diverse environments, M3ED offers a comprehensive and realistic foundation for studying event-driven perception. Its mixture of high-speed motion, complex illumination, and irregular camera trajectories provides a challenging setting that enables robust evaluation of temporal fusion, motion forecasting, and multimodal reasoning algorithms in real-world robotic scenarios.

\subsubsection{{DSEC Dataset}}
The DSEC dataset is a large-scale multimodal driving benchmark collected using a tightly synchronized stereo sensor rig composed of two Prophesee Gen3.1 event cameras ($640\times480$) and two FLIR Blackfly S RGB cameras operating at 20\,Hz. All four sensors are factory-calibrated, geometrically rectified, and time-synchronized at the microsecond level, enabling pixel-accurate fusion of RGB frames and asynchronous events. The recordings span multiple Swiss cities, including Zurich, Thun, and Interlaken, and cover a wide variety of driving environments such as busy downtown streets, narrow suburban lanes, mountain tunnels, and high-speed highway segments. In total, DSEC contains over one hour of real-world driving with typical sequence durations of 2–8 minutes.

Compared with earlier event-camera driving datasets, DSEC emphasizes geometric accuracy and multimodal alignment. The dataset provides stereo rectification, precise extrinsics, and high-fidelity rolling-shutter compensation for RGB cameras, ensuring consistency between modalities even under high-speed motion. The event stream captures fine-grained temporal structure with microsecond resolution, making DSEC well suited for studying motion blur suppression, optical flow estimation, and low-latency perception in dynamic urban scenes.

Building on DSEC, DAGr~\cite{gehrig2024dagr} introduces \textbf{DSEC-Detection}, an enhanced object detection benchmark that augments the original sequences with dense, spatially aligned \textbf{2D bounding-box} annotations. A depth-free geometric warping procedure is applied to map RGB frames into the event-camera viewpoint, reducing residual parallax to within 6\,px across the image plane. Object trajectories are first generated using QDTrack and subsequently verified and corrected by human annotators. Additional pedestrian-centric scenes are introduced to mitigate the category imbalance inherent in the original dataset.

DSEC-Detection offers high-quality annotations for eight categories across long, continuous sequences. Notably, approximately one-third of the annotated frames contain object appearance or disappearance induced by fast ego-motion, dynamic occlusions, or sudden viewpoint changes, making DSEC-Detection a challenging benchmark for event-driven detection and motion-aware fusion. Together, DSEC and DSEC-Detection provide a geometrically precise, temporally rich, and motion-intensive foundation for evaluating asynchronous perception and event–frame alignment algorithms in real-world driving conditions.

\begin{table*}[t]
\centering
\caption{
Statistics of the \eventdrive dataset across tasks, sequences, and annotated frames.
}
\resizebox{\linewidth}{!}{
\begin{tabular}{ll|rrr|rrr|rrr}
\toprule

& & \multicolumn{3}{c|}{\textbf{Train}} & \multicolumn{3}{c|}{\textbf{Test}} & \multicolumn{3}{c}{\textbf{Hard}}
\\
\textbf{Task} & \textbf{Source} 
& \textbf{\#Seq} & \textbf{\#Frame} & \textbf{\#Data} 
& \textbf{\#Seq}  & \textbf{\#Frame}  & \textbf{\#Data}
& \textbf{\#Seq}  & \textbf{\#Frame}  & \textbf{\#Data} \\
\midrule\midrule

\multirow{3}{*}{
\includegraphics[width=0.022\linewidth]{icons/perception.png}~\textbf{Perception}
}
& DSEC \cite{gehrig2021dsec} 
& $47$ & $4{,}433$ & $53{,}196$ 
& $12$ & $1{,}069$ & $12{,}828$
& $1$ & $65$ & $780$ \\

& M3ED \cite{chaney2023m3ed} 
& $3$ & $5{,}440$ & $65{,}280$ 
& $2$ & $2{,}678$ & $32{,}136$
& $1$ & $1{,}283$ & $15{,}396$ \\

& PKU \cite{li2023sodformer}
& $31$ & $3{,}905$ & $46{,}860$ 
& $10$ & $950$ & $11{,}400$
& $3$ & $330$ & $3{,}960$ \\
\midrule

\includegraphics[width=0.022\linewidth]{icons/understanding.png}~\textbf{Understanding}
& DSEC \cite{gehrig2021dsec}
& $47$ & $4{,}433$ & $91{,}716$ 
& $12$ & $1{,}069$ & $28{,}848$
& $1$ & $65$ & $1{,}812$ \\

\includegraphics[width=0.022\linewidth]{icons/prediction.png}~\textbf{Prediction}
& M3ED \cite{chaney2023m3ed}
& $3$ & $3{,}892$ & $7{,}784$
& $2$ & $1{,}405$ & $2{,}810$
& $1$ & $985$ & $1{,}970$ \\

\includegraphics[width=0.022\linewidth]{icons/planning.png}~\textbf{Planning}
& M3ED \cite{chaney2023m3ed}
& $3$ & $15{,}430$ & $46{,}290$
& $2$ & $9{,}842$ & $29{,}526$
& $1$ & $6{,}317$ & $18{,}951$ \\
\midrule

\rowcolor{gray!20}
\multicolumn{2}{l}{\textbf{Grand Total Samples}} 
& \multicolumn{9}{c}{$\mathbf{471{,}543}$} \\
\bottomrule
\end{tabular}
}
\label{tab:eventdrive_stats}
\end{table*}

\subsubsection{{PKU-DAVIS-SOD Dataset}}
The PKU-DAVIS-SOD dataset is a large-scale multimodal benchmark designed for real-world object detection under challenging illumination and motion conditions. It is collected using a DAVIS346 sensor that streams high-frequency DVS events together with global-shutter RGB frames ($346\times260$ at 25\,Hz). The combined sensing pipeline enables robust annotation even in scenarios that traditionally hinder frame-based perception, such as extreme motion blur, low-light urban environments, and sudden illumination transitions. Overall, the dataset consists of \textbf{220 continuous driving sequences}, offering long, uninterrupted event–frame streams rather than isolated clips.

A distinguishing feature of PKU-DAVIS-SOD is its dense, frame-wise bounding-box supervision. The benchmark provides \textbf{276k labeled timestamps} and \textbf{1.08M manually annotated bounding boxes} across three object categories: \textbf{cars}, \textbf{pedestrians}, and \textbf{two-wheelers}. To ensure accuracy under fast motion or low-light conditions, annotations are supported by event-reconstructed grayscale images that preserve structural edges where RGB frames may fail. The dataset is split into 671.3k / 194.7k / 214.1k bounding boxes for train/val/test, maintaining balanced distributions of object sizes (small, medium, large) within each split.

The sequences capture a wide spectrum of urban driving conditions. Approximately 92\% of data is recorded under normal illumination, while 8\% focuses on low-light scenes including dimly lit roads, urban night drives, and shadowed intersections. Motion patterns also vary significantly: normal-speed trajectories account for 87\% of frames, whereas 13\% correspond to high-speed or motion-blur scenarios where events provide crucial temporal detail unavailable in RGB alone. These diverse conditions make PKU-DAVIS-SOD a valuable testbed for studying multimodal robustness and motion-induced edge cases.

Beyond object detection, PKU-DAVIS-SOD enables research in asynchronous fusion, temporal reasoning, and motion-aware event processing. Its long continuous streams, high annotation density, and broad coverage of real-world driving scenarios offer a strong foundation for evaluating fine-grained perception algorithms that must handle illumination variability, high-speed dynamics, and sparse-but-informative event modality signals.

These three datasets collectively provide (1) heterogeneous viewpoints and motion regimes (ground, aerial, quadruped), (2) multiple sensor generations (Gen3.1, Gen4 event cameras), (3) both short and long temporal horizons, and (4) dense multimodal annotations spanning 2D/3D detection, segmentation, optical flow, and odometry. This diversity allows \eventdrive to unify perception, understanding, prediction, and planning within a single event-frame language model while maintaining broad generalization across motion scales, illumination conditions, and real-world driving scenarios.

\subsection{Dataset Construction Strategy}

To construct a unified multimodal benchmark that spans perception, understanding, prediction, and planning, we curate data from three complementary event-frame datasets: \textbf{DSEC}, \textbf{M3ED}, and \textbf{PKU-DAVIS-SOD}. These datasets jointly provide high-quality RGB-event pairs, dense 2D annotations, LiDAR-supported 3D motion labels, and long continuous trajectories under diverse daytime, nighttime, and motion-blur conditions. However, their sensing configurations, temporal resolutions, and scene distributions differ substantially. We therefore apply task-specific filtering rules and consistent sequence-level splits to form a balanced, coherent corpus for \eventdrive.

\noindent \textbf{\perception Split Construction.}
Perception focuses on object and scene recognition and relies only on synchronized image-event pairs. We include all three source datasets while enforcing dataset-specific sampling procedures to ensure temporal diversity and avoid redundancy. For \textbf{DSEC}, we adopt the official train/validation/test partition. Because consecutive RGB frames change minimally at 20\,Hz, we subsample every eighth frame to reduce duplication while maintaining temporal coverage. The sequence \textit{zurich\_city\_12\_a}, a nearly pitch-black urban drive with extremely low illumination, is designated as the hard split. For \textbf{M3ED}, we retain only ground-vehicle recordings, discarding UAV and quadruped runs due to incompatible motion characteristics, along with extremely short clips and parking-lot-only trajectories. The final selection includes three daytime sequences (\textit{urban\_day\_city\_hall}, \textit{urban\_day\_rittenhouse}, \textit{urban\_day\_ucity\_small\_loop}) and three nighttime sequences (\textit{urban\_night\_city\_hall}, \textit{urban\_night\_rittenhouse}, \textit{urban\_night\_ucity\_small\_loop}). We follow a symmetric day/night protocol in which \textit{urban\_day\_rittenhouse} and \textit{urban\_night\_rittenhouse} form the test set, while the long sequence \textit{urban\_night\_ucity\_small\_loop} is used as the hard split. For \textbf{PKU-DAVIS-SOD}, which has lower resolution and noisier appearance, we rely on the official validation split containing 28 normal, 10 motion-blur, and 6 low-light sequences. We select eight normal sequences (21-28), one low-light, and one motion-blur sequence for testing, and designate one low-light and two motion-blur sequences as hard cases. This selection preserves a balanced distribution of illumination and motion-blur conditions.

\noindent \textbf{\understanding Split Construction.}
Scene understanding requires high-quality 2D bounding boxes. We therefore use DSEC exclusively, leveraging the dense annotations of DSEC-DET together with its high-resolution stereo views. We follow the same sequence-level split and 8-frame subsampling strategy used in perception, while additionally ensuring that each selected frame contains at least one valid instance annotation. This produces a reliable, densely annotated corpus for event-driven scene understanding.

\noindent \textbf{\prediction Split Construction.}
Trajectory prediction depends on accurate agent motion and ego-motion annotations, which M3ED provides through synchronized Ouster LiDAR sweeps, globally registered odometry, and 3D instance tracking from \cite{liang2025perspective}. We inherit the perception split for sequence selection and further refine it by removing segments with unstable ground truth, including those with severely noisy 3D boxes, prolonged occlusions, or substantial pose drift. The resulting subset offers consistent agent trajectories and diverse velocity profiles suitable for motion forecasting.

\noindent \textbf{\planning Split Construction.}
Planning supervision requires future ego-vehicle trajectories, which are also provided only by M3ED through FasterLIO-based odometry. We adopt the same train/test/hard split as prediction to ensure consistency across the motion-centric tasks. For each retained sequence, we generate high-level driving intents (speed and path) together with 5-second future waypoint sequences sampled at 0.5-second intervals.

Across all four reasoning tasks, \eventdrive integrates 471{,}543 multimodal samples drawn from three datasets. The corpus spans urban daytime and nighttime driving, tunnel environments, motion-blur episodes, rapid rotations, and extreme low-light conditions. Compared with previous event-language datasets, our dataset provides the largest event-frame-language corpus for driving, the only dataset that covers all four reasoning levels in a single benchmark, diverse cross-domain and cross-illumination conditions, and long-horizon trajectory supervision necessary for learning event-driven prediction and planning. \cref{tab:eventdrive_stats} summarizes the curated splits.

\begin{figure}[t]
    \centering
    \includegraphics[width=\linewidth]{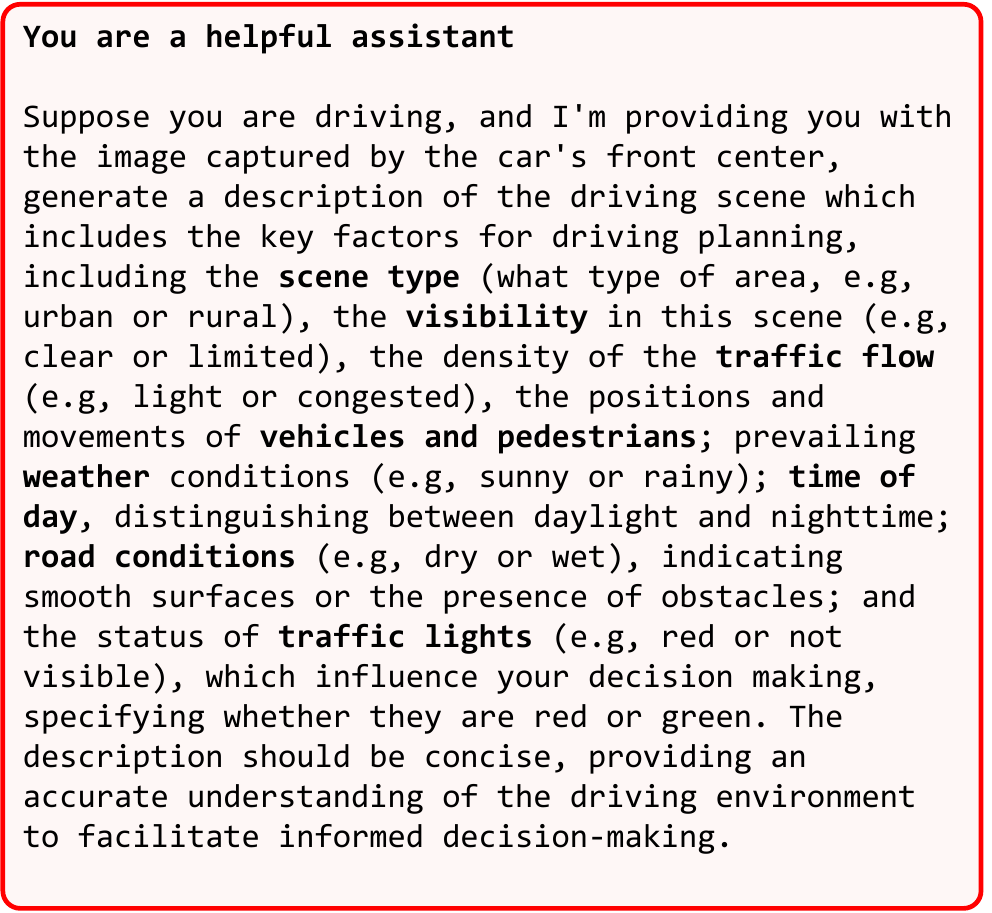}
    \caption{
    Prompt used to generate a scene caption capturing six essential attributes of the driving scene.
    }
    \label{fig:perception_caption}
\end{figure}

\subsection{Annotation Pipeline}

Beyond curating and splitting raw sequences, \eventdrive requires transforming heterogeneous sensor data into unified, language-driven supervision suitable for multimodal reasoning. Each task demands different forms of annotation, ranging from structured scene attributes and object-centric queries to agent trajectories and ego-motion intentions. To achieve this, we design a modular annotation pipeline that combines rule-based filtering, trajectory processing, and large vision-language model prompting.

The pipeline converts raw RGB-event pairs, 2D/3D detections, and LiDAR-supported trajectories into standardized instruction-following QA formats. For semantics-driven tasks (perception and understanding), we employ caption generation and caption-to-QA transformation with controlled distractors. For motion-centric tasks (prediction and planning), we process agent and ego trajectories in the ego frame and derive high-level speed and path intents via kinematic rules. All annotations are formatted into consistent two-turn Qwen-style conversations, ensuring that every modality, including event, image, and text, contributes a coherent supervisory signal for training unified event-driven VLMs.

\subsubsection{\perception Annotation}
To construct perception-level annotations, we design a semi-automated pipeline that transforms raw driving images into six structured multiple-choice QA pairs representing key scene attributes. The full process consists of three stages: scene caption generation, caption-to-QA conversion, and final formatting with controlled distractors.

\begin{figure}[t]
    \centering
    \includegraphics[width=\linewidth]{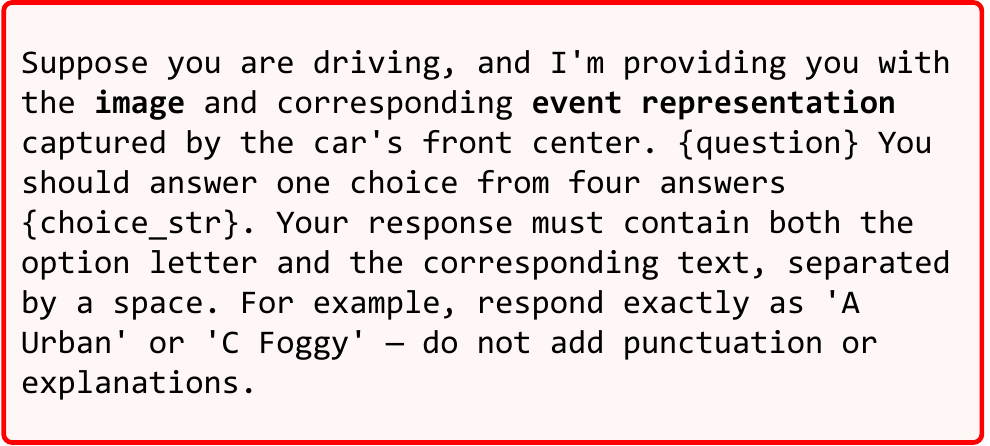}
    \caption{
    Prompt used to assess the model’s \perception capability in driving scenes.
    }
    \label{fig:infer_perception}
\end{figure}

\noindent\textcolor{frame_red}{{\footnotesize$\blacksquare$}}~\textbf{\textit{Stage 1: Scene Caption Generation.}}
We begin by generating a rich and driving-oriented description for each RGB frame. A high-capacity vision–language model (Qwen3-VL-30B-A3B-Instruct \cite{qwen3}) receives the front-view image together with a prompt, as shown in \cref{fig:perception_caption}, that instructs it to summarize all scene factors essential for driving. This includes environmental type (e.g., urban, suburban), visibility, traffic density, agent motion, weather, time of day, road surface condition, and traffic-light state. Images are processed in batches for efficiency. These captions serve as a structured semantic foundation for subsequent QA construction.

\begin{figure*}[t]
    \centering
    \includegraphics[width=\linewidth]{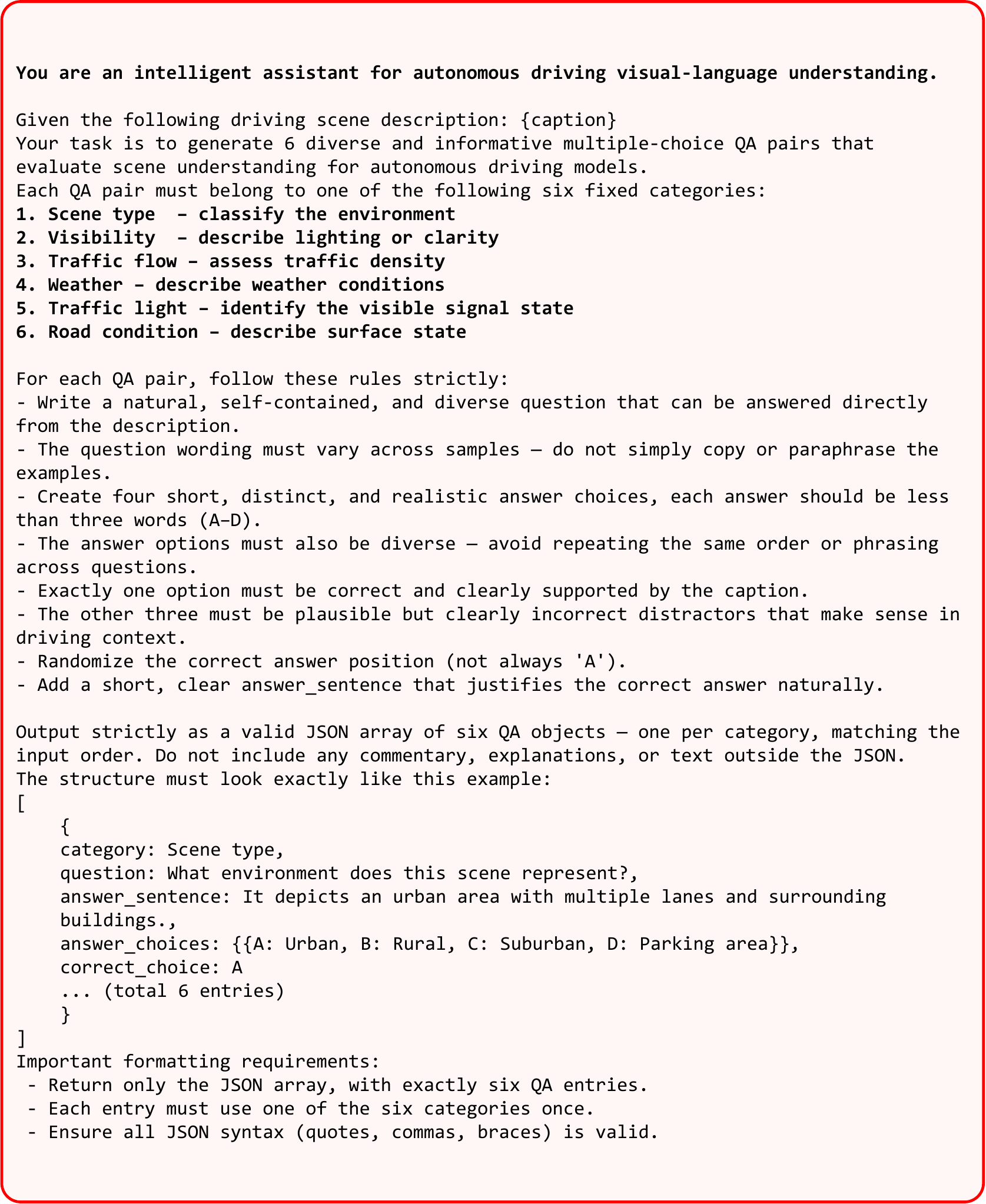}
    \caption{
    Prompt used to generate QA pairs from a scene caption that encodes six essential attributes of the driving scene.
    }
    \label{fig:perception_qa}
\end{figure*}

\noindent\textcolor{frame_red}{{\footnotesize$\blacksquare$}}~\textbf{\textit{Stage 2: Caption-to-QA Generation.}}
Given the caption for each image, we use Qwen2.5-VL-7B-Instruct to transform the description into six perception QA pairs, one for each predefined category:
\textit{$^1$Scene type}, \textit{$^2$Visibility}, \textit{$^3$Traffic flow}, \textit{$^4$Weather}, \textit{$^5$Traffic light}, and \textit{$^6$Road condition}.
The prompt (shown in \cref{fig:perception_qa}) instructs the model to:
(1) produce natural, self-contained questions grounded in the caption,  
(2) offer four candidate answers (A–D) with exactly one correct option,  
(3) ensure wording diversity across samples,  
(4) avoid repetitive patterns in answer ordering, and  
(5) provide a short justification sentence tied to the caption.
The model outputs a strict JSON array with six entries. Outputs that fail JSON parsing are kept for later inspection but excluded from final training data.

\begin{figure}[t]
    \centering
    \includegraphics[width=\linewidth]{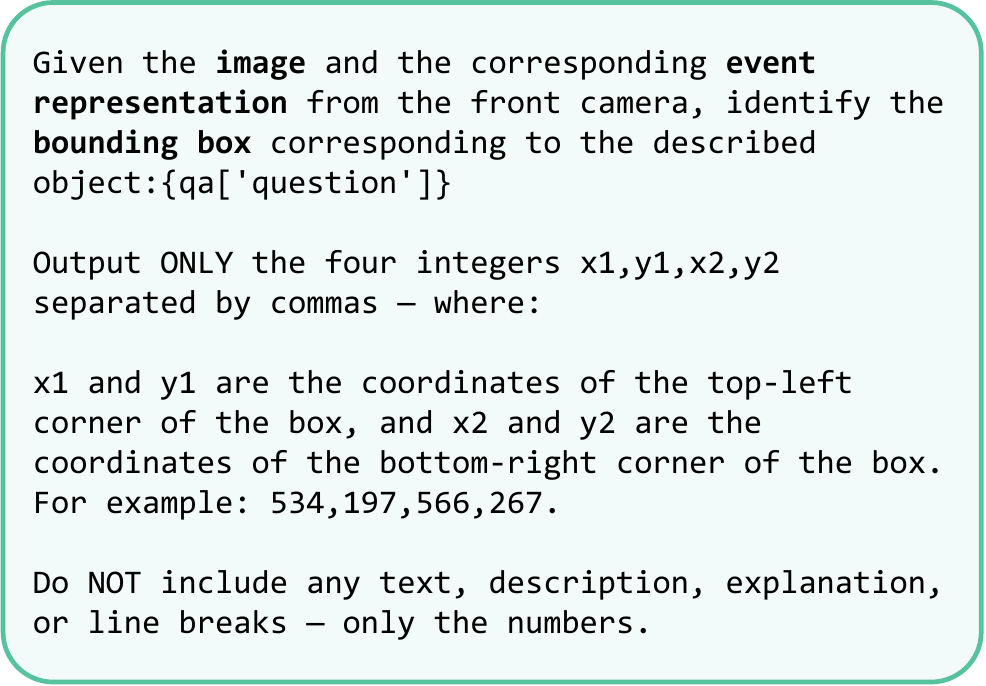}
    \caption{
    Prompt used to assess the model’s \textbf{grounding} capability in driving scenes.
    }
    \label{fig:infer_grounding}
\end{figure}

\begin{figure}[t]
    \centering
    \includegraphics[width=\linewidth]{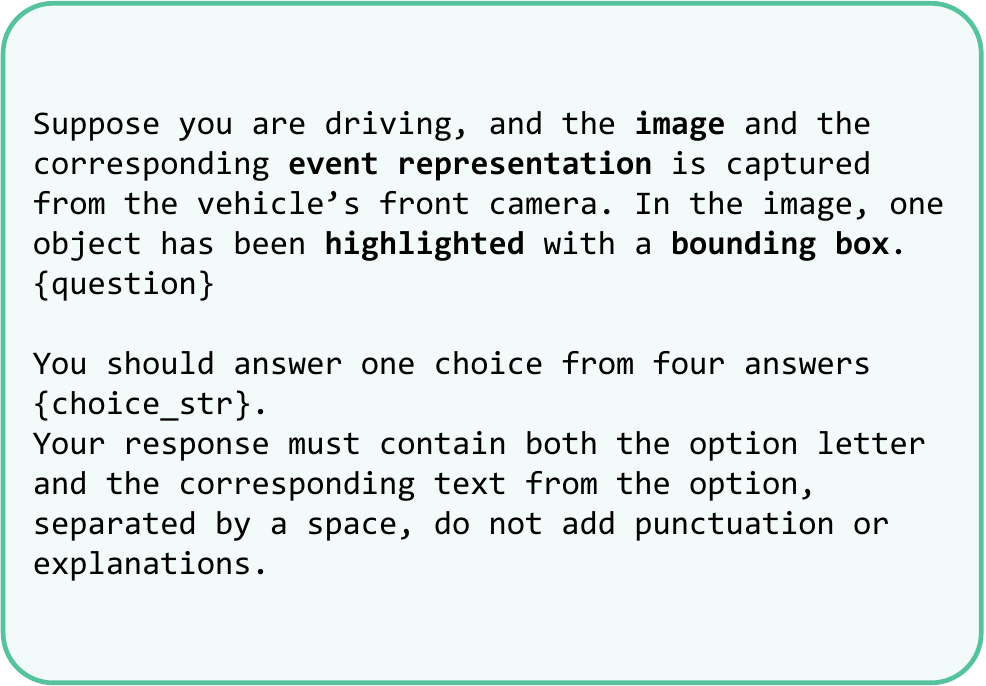}
    \caption{
    Prompt used to assess the model’s \understanding capability in driving scenes.
    }
    \label{fig:infer_understanding}
\end{figure}

\noindent\textcolor{frame_red}{{\footnotesize$\blacksquare$}}~\textbf{\textit{Stage 3: Multiple-Choice Construction and Formatting.}}
The generated QA objects are then converted into the final format used by our training pipeline. 
Although the LLM provides candidate answers, we replace all options with a controlled set of answer choices for consistency across the dataset. 
For each attribute category, we maintain a manually curated distractor pool covering realistic but incorrect alternatives relevant to driving scenes (e.g., various weather types, road-surface states, visibility conditions, or traffic-light statuses). For each QA instance, we:
(1) extract the correct answer from the LLM output,  
(2) sample three plausible distractors from the category’s distractor pool,  
(3) shuffle the four options to randomize the correct label position,  
(4) construct a two-turn conversation in the Qwen training format. The human turn contains the driving context, the question, and the four answer options, while the assistant turn outputs the ground-truth choice in the required ``letter text'' format (e.g., ``B Low light'').

Once the QA pairs are constructed, we apply the prompting template illustrated in \cref{fig:infer_perception} to perform inference over RGB-event inputs and generate the final perception results used for evaluation.

This pipeline combines generative models with controlled distractor sampling and strict formatting rules to produce high-quality, diverse, and semantically grounded perception annotations.  
By leveraging LLM-generated captions and ensuring category-level consistency through curated distractor pools, the resulting QA dataset offers a robust benchmark for evaluating scene-level understanding under both RGB and event-based sensing.

\subsubsection{\understanding Annotation}
The annotation pipeline for object-level understanding follows a structured, multi-stage process that combines LLM-based object captioning, rule-driven QA construction, bounding-box candidate generation, and final conversion into Qwen-style supervision. This dimension requires fine-grained descriptions for individual objects, strict uniqueness validation, and multiple task types (appearance, grounding, relations, etc.).

\begin{figure}[t]
    \centering
    \includegraphics[width=\linewidth]{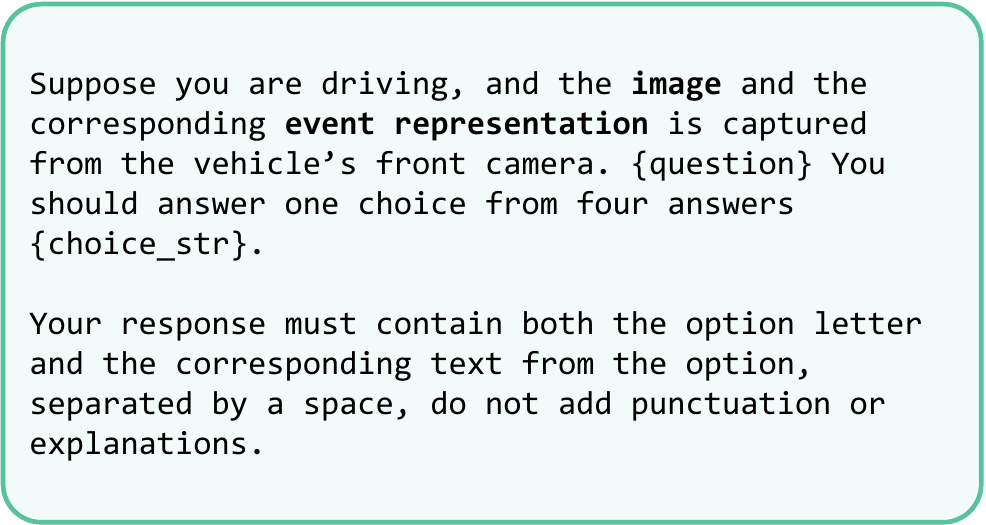}
    \caption{
    Prompt used to assess the model’s \textbf{object-awareness} capability in driving scenes.
    }
    \label{fig:infer_object}
\end{figure}

\begin{figure*}[t]
    \centering
    \includegraphics[width=\linewidth]{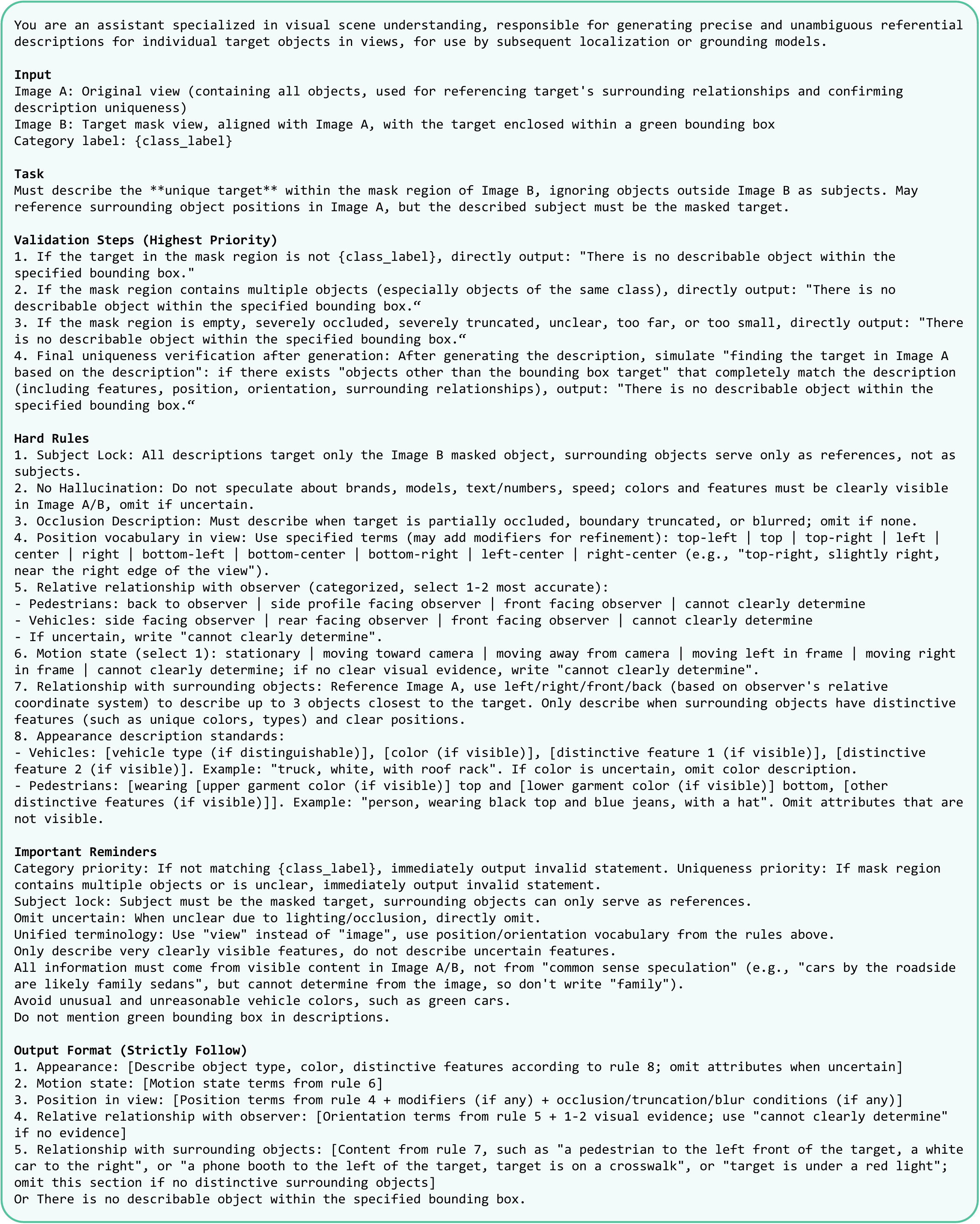}
    \caption{
    Prompt used to generate an object caption capturing essential object attributes of the driving scene.
    }
    \label{fig:understanding_caption}
\end{figure*}

\noindent\textcolor{event_green}{{\footnotesize$\blacksquare$}}~\textbf{\textit{Stage 1: Generating Object-Level Referential Descriptions}}
For each ground-truth object, we provide an LLM with (1) the full-frame scene image, (2) a masked view where the target is enclosed by a bounding box, and (3) the object’s class label. A specialized prompt, as shown in \cref{fig:understanding_caption}, instructs the model to first perform a series of validity checks, ensuring that the boxed region contains exactly one object of the correct class and that the target is sufficiently visible, unambiguous, and not overly occluded or truncated. If any of these conditions fail, the model must output the rejection statement: 
\emph{``There is no describable object within the specified bounding box.''}

If the region is valid, the model generates a structured referential description following a constrained five-part template, covering \textit{$^1$appearance}, \textit{$^2$motion state}, \textit{$^3$position in the view}, \textit{$^4$relation to the viewer}, and \textit{$^5$relation to surrounding objects}. The prompt strictly prohibits hallucination, speculation, uncertain attributes, or treating nearby objects as subjects. All spatial terminology follows a fixed vocabulary (e.g., ``top-right'', ``center-left''), and descriptions must uniquely identify the boxed object without mentioning the bounding box itself. This produces consistent, grounded, and disambiguated object-level captions.

\begin{figure*}[t]
    \centering
    \includegraphics[width=\linewidth]{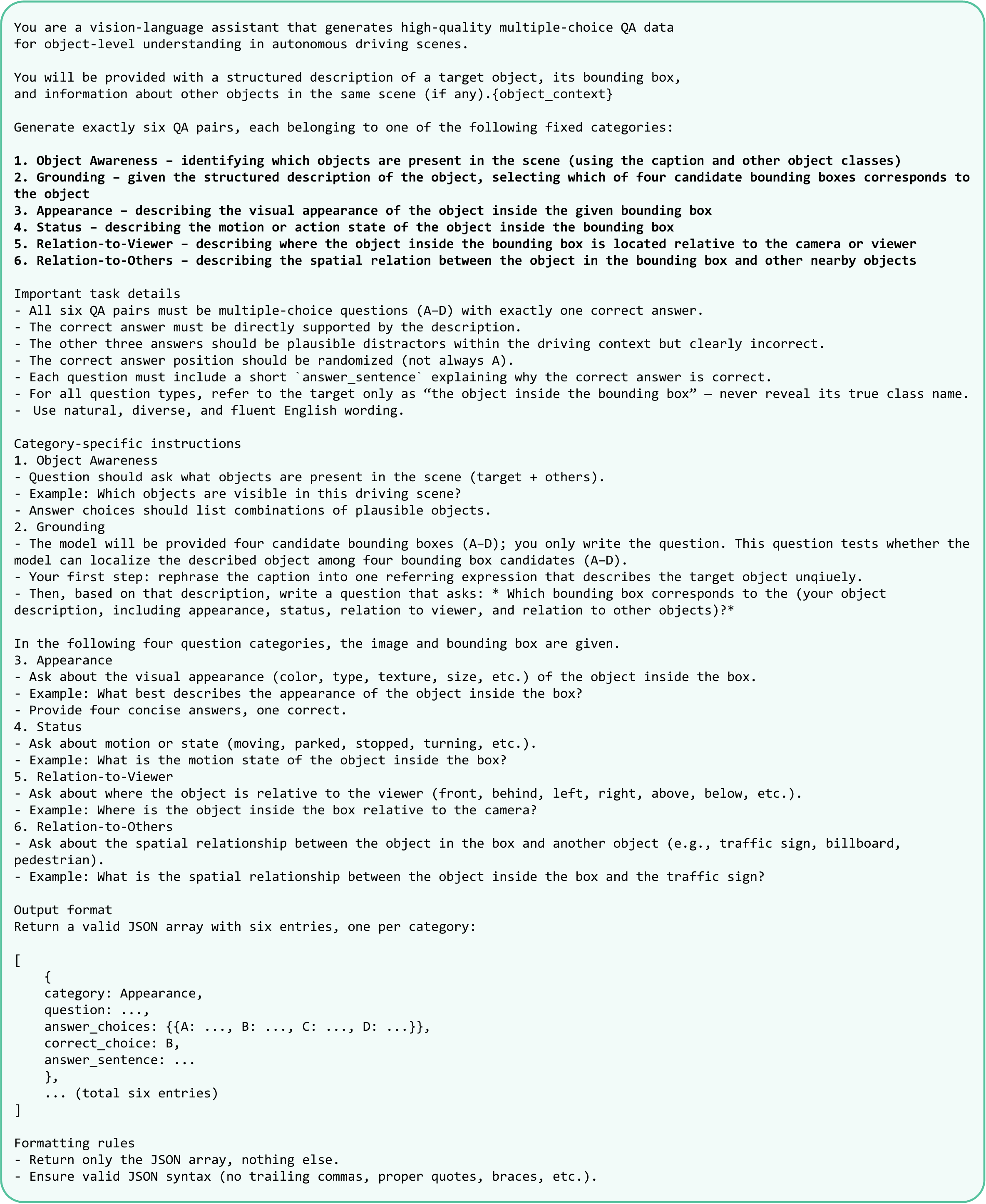}
    \caption{
    Prompt used to generate QA pairs from an object caption that encodes essential object attributes in the driving scene.
    }
    \label{fig:understanding_qa}
\end{figure*}

\noindent\textcolor{event_green}{{\footnotesize$\blacksquare$}}~\textbf{\textit{Stage 2: Converting Captions into Six QA Categories.}}
Each structured description is then passed to a second LLM, which generates exactly six multiple-choice QA pairs, one for each object-understanding dimension: \textit{$^1$object awareness}, \textit{$^2$grounding}, \textit{$^3$appearance}, \textit{$^4$status}, \textit{$^5$relation-to-viewer}, and \textit{$^6$relation-to-others}. The prompt, as shown in \cref{fig:understanding_qa}, enforces uniform formatting: four answer options (A–D) with exactly one correct choice, plausible distractors, randomized correct-answer position, and a short justification sentence. For grounding, the model produces a referring expression but does not invent coordinates, as bounding-box candidates are injected later. The LLM returns a clean JSON array containing the six QA entries.

\noindent\textcolor{event_green}{{\footnotesize$\blacksquare$}}~\textbf{\textit{Stage 3: Preparing Bounding-Box Candidates for Grounding.}}
For grounding questions, we construct four candidate bounding boxes by combining the ground-truth box with up to three distractors drawn from other objects in the same frame. If fewer distractors are available, additional jittered boxes with randomized size and location (remaining within image bounds) are synthesized. The four candidates are then shuffled and reassigned labels, and the correct label is recorded. This ensures reliable localization supervision while preventing positional bias.

\noindent\textcolor{event_green}{{\footnotesize$\blacksquare$}}~\textbf{\textit{Stage 4: Converting QA Items into Qwen Training Format.}}
Finally, each QA pair is converted into a two-turn dialogue compatible with Qwen-style multimodal training. For \textit{non-grounding} tasks, the human turn introduces the driving context, highlights that an object is boxed in the image, presents the question and four answer choices, and instructs the model to answer in the exact ``letter~text'' format (e.g., ``C Moving left''). The assistant turns outputs the correct label and answer. \textit{Grounding} questions follow a stricter interface: the human turn requests only the four integers $(x, y, w, h)$ corresponding to the selected box, with no additional text, where $(x, y)$ denotes the top-left corner and $(w, h)$ denotes the box width and height. The assistant turn outputs the correct coordinates. Each sample stores the image path, event path, QA category, and the ground-truth box for later evaluation. All QA pairs are flattened so that each becomes an independent training example. Once the QA pairs are generated, we use the prompting templates shown in \cref{fig:infer_object} for object-awareness questions, \cref{fig:infer_understanding} for attribute-based understanding that requires bounding-box inputs, and \cref{fig:infer_grounding} for grounding. These templates are applied to RGB–event inputs to obtain the final understanding results used for evaluation.

This four-stage pipeline transforms raw bounding-box annotations into rich object-level supervision: validated referential descriptions, six structured QA pairs per instance, standardized grounding candidates, and fully formatted Qwen dialogue samples. The result is a consistent and scalable object-understanding dataset covering appearance, localization, motion, and relational reasoning across diverse driving scenes.

\subsubsection{\prediction Annotation}
The prediction annotation pipeline focuses on generating supervisory labels for short-horizon motion prediction of dynamic agents in driving scenes. Unlike perception or understanding, which rely primarily on static visual cues, prediction requires integrating multi-frame geometric information, ego-vehicle pose, and local object motion. The pipeline proceeds through three major stages: (1) deriving object trajectories in the ego coordinate frame, (2) converting physical motion into discrete intent labels, and (3) constructing image-conditioned QA samples for model training and evaluation.

\noindent\textcolor{fusion_blue}{{\footnotesize$\blacksquare$}}~\textbf{\textit{Stage 1: Ego-Frame Trajectory Construction.}}
For each video sequence, we load the LiDAR-based 3D detection and tracking results \cite{liang2025perspective}, obtaining the world-frame positions of surrounding dynamic agents at $10$\,Hz.  
For each tracked agent, its center position is transformed into the ego-vehicle coordinate system at every timestamp, producing a short-term trajectory
\[
\mathcal{Q} = \{\mathbf{q}_t = (x_t,\, y_t,\, z_t)\}_{t=0}^{10},
\]
where $y$ denotes the forward axis, $x$ the lateral axis, and the sampling interval is $\Delta t = 0.1$\,s.

To analyze motion relative to the ego vehicle, we normalize the trajectory by subtracting the initial position:
\[
\tilde{\mathbf{q}}_t = \mathbf{q}_t - \mathbf{q}_0 .
\]
This removes global-location bias and ensures that the resulting displacement reflects the agent’s motion purely from the ego-vehicle’s perspective.

\noindent\textcolor{fusion_blue}{{\footnotesize$\blacksquare$}}~\textbf{\textit{Stage 2: Deriving Speed and Path Intent Labels.}}
Given the $1$-second relative trajectory $\{\tilde{\mathbf{q}}_t\}$, we infer the agent's semantic motion intent along two axes: \emph{speed intent} and \emph{path intent}.

\noindent\textit{1) Speed Intent.}
Velocity is estimated using finite differences, following \cite{chi2025impromptuvla},
\[
v_t = \frac{\lVert \tilde{\mathbf{q}}_{t+1} - \tilde{\mathbf{q}}_{t} \rVert}{\Delta t},
\]
and intent is determined by thresholding the velocity change across the trajectory:
\begin{itemize}
    \item \texttt{STOP}: all velocities remain below a small threshold;
    \item \texttt{ACCELERATE}: final velocity significantly exceeds the initial velocity;
    \item \texttt{DECELERATE}: final velocity drops noticeably below the initial velocity;
    \item \texttt{KEEP}: velocity remains approximately constant.
\end{itemize}
This rule-based design follows conventions in autonomous driving motion-forecasting benchmarks.

\noindent\textit{2) Path Intent.}
Directional motion is determined from the final displacement:
\[
\Delta x = x^{\text{future}}_{10}, \qquad
\Delta y = y^{\text{future}}_{10}.
\]
The path intent is assigned by comparing lateral drift against a fixed threshold while ensuring that longitudinal displacement indicates stable forward motion:
\begin{itemize}
    \item \texttt{LEFT}: $\Delta x$ exceeds a positive threshold, indicating object moving toward the ego-vehicle's left side;
    \item \texttt{RIGHT}: $\Delta x$ is below a negative threshold, indicating a rightward drift;
    \item \texttt{STRAIGHT}: longitudinal displacement dominates, i.e., $|\Delta x|$ is small relative to $|\Delta y|$;
    \item \texttt{UNKNOWN}: displacement is too small or inconsistent to reliably infer direction.
\end{itemize}

\begin{figure}[t]
    \centering
    \includegraphics[width=\linewidth]{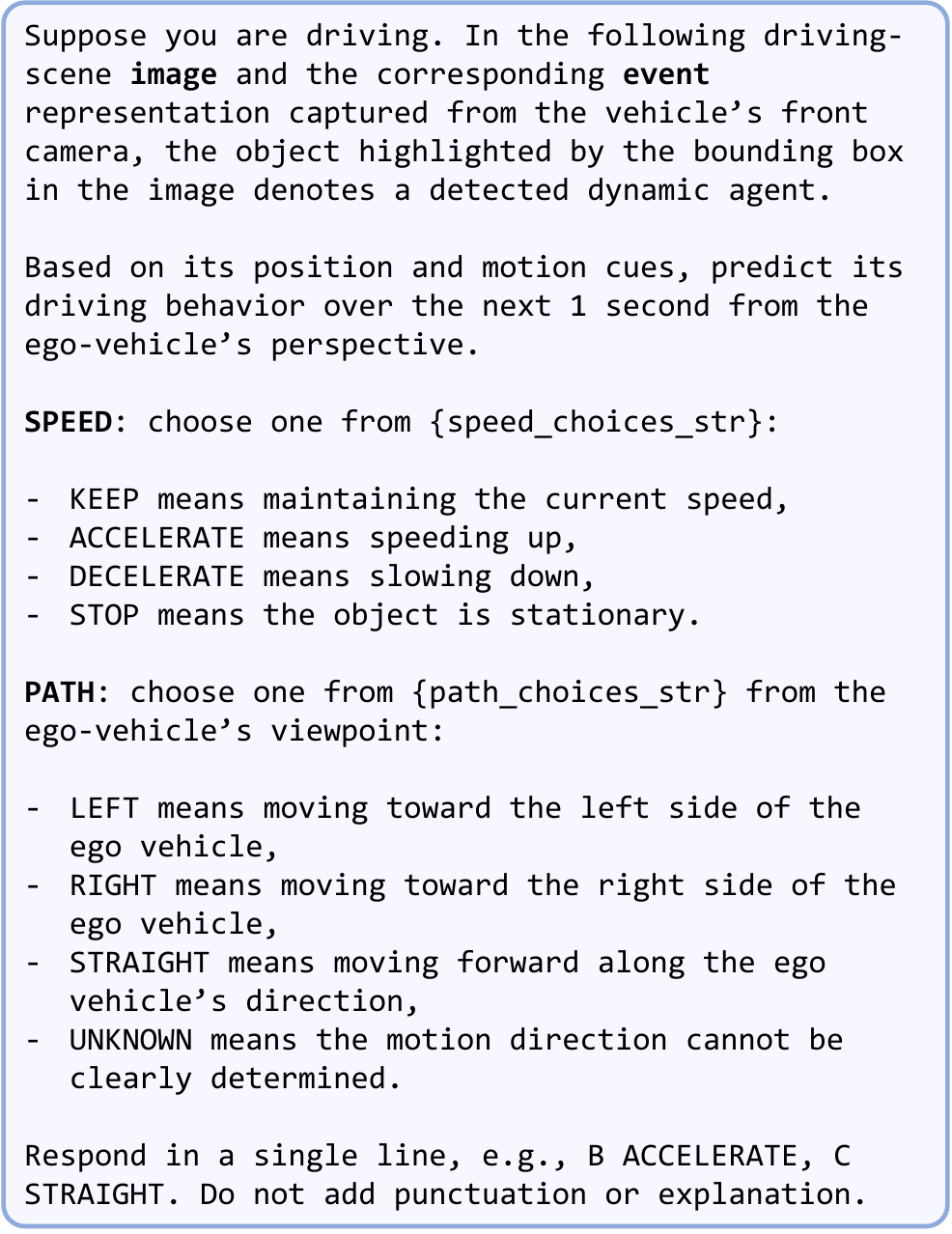}
    \caption{
    Prompt used to assess the model’s \prediction capability in driving scenes.
    }
    \label{fig:infer_prediction}
\end{figure}

\noindent\textcolor{fusion_blue}{{\footnotesize$\blacksquare$}}~\textbf{\textit{Stage 3: Building Prediction QA Samples.}}
Each tracked agent contributes one prediction QA instance.  
Given the RGB frame and event stream with the agent highlighted by a bounding box, the model must output the two-token intent classification
\[
\texttt{<SPEED>, <PATH>}
\]
(e.g., ``\texttt{DECELERATE, LEFT}'' or ``\texttt{STOP, UNKNOWN}'').  
This yields a unified training format for multimodal intent prediction from RGB--event inputs. Once the QA pairs are constructed, we apply the prompting template shown in \cref{fig:infer_prediction} to perform RGB-event inference, producing the final prediction results that are used for evaluation.

The prediction annotation pipeline transforms 3D-detected agent trajectories into semantically meaningful short-horizon motion intents and pairs them with image–event evidence through structured QA formatting. By grounding motion labels in ego-centric geometry and enforcing standardized multiple-choice output, the pipeline provides a clean and scalable supervision scheme that unifies physical trajectory prediction with multimodal language-model training.

\subsubsection{\planning Annotation}

The planning annotation process generates two types of supervision:
(1) high-level driving intent (\emph{speed} and \emph{path} decisions), and
(2) low-level ego-trajectory forecasting.
The pipeline consists of four main stages, combining ego-pose extraction, ego-frame trajectory normalization, rule-based intent derivation, and Qwen-style QA construction.

\begin{figure}[t]
    \centering
    \includegraphics[width=\linewidth]{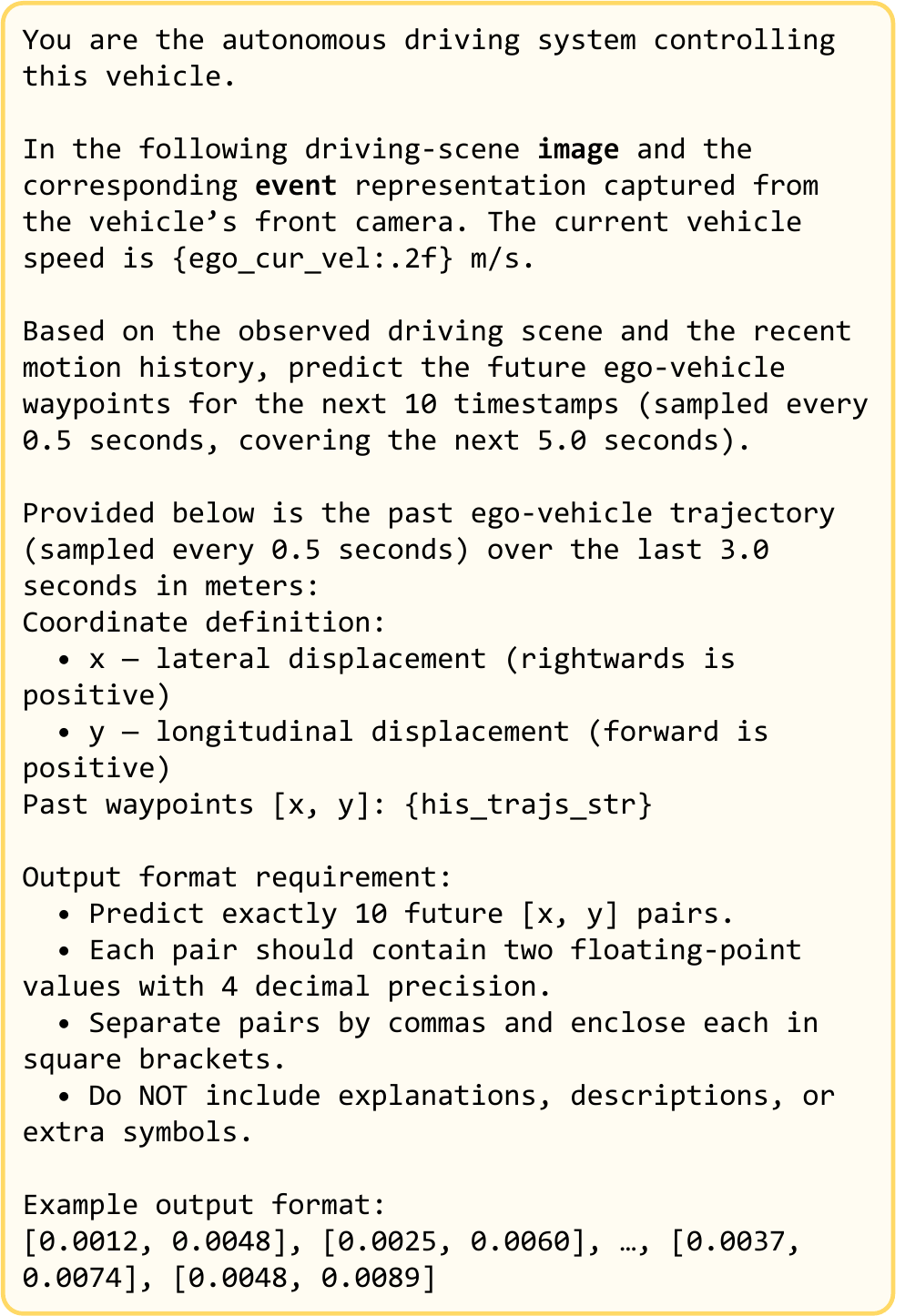}
    \caption{
    Prompt used to assess the model’s ego waypoint \planning capability in driving scenes.
    }
    \label{fig:infer_planning_waypoint}
\end{figure}

\begin{figure}[t]
    \centering
    \includegraphics[width=\linewidth]{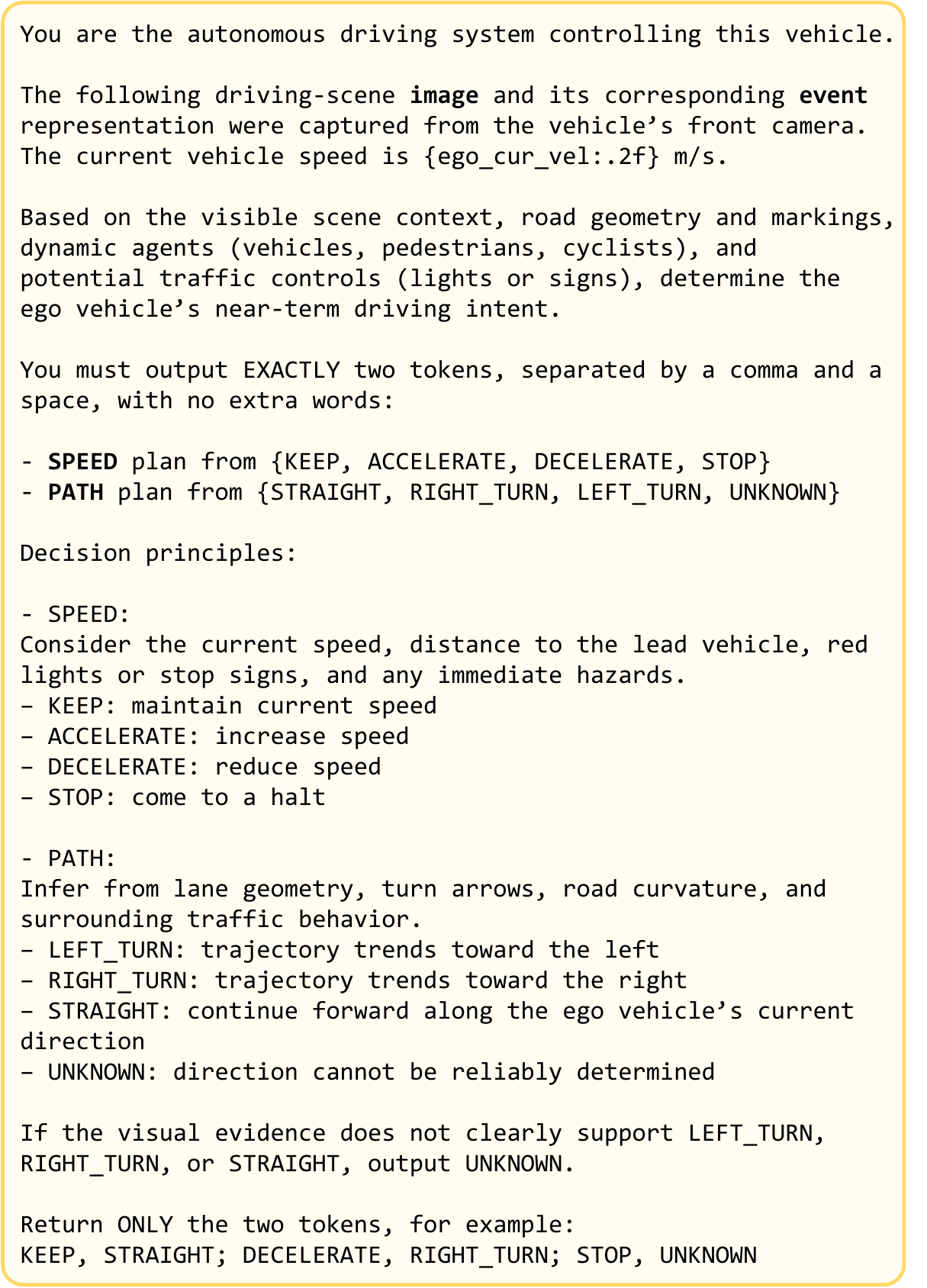}
    \caption{
    Prompt used to assess the model’s ego path and speed \planning capability in driving scenes.
    }
    \label{fig:infer_planning_intent}
\end{figure}

\noindent\textcolor{planning_gold}{{\footnotesize$\blacksquare$}}~\textbf{\textit{Stage 1: Ego-Frame Trajectory Construction.}} For each sequence, we first load the vehicle's global poses from the ground-truth pose logs, where each record contains a timestamp, position $(t,\,x,\,y,\,z)$ and orientation represented as a quaternion.  
Given the image timestamp, the nearest pose is identified, and a temporal window of the past $3$ seconds and future $5$ seconds is collected around this index.  Let the global trajectory be:
\[
\mathcal{P} = \{ \mathbf{p}_t = (x_t, y_t, z_t) \}_{t = -T_\text{past}}^{t = T_\text{future}} .
\]
To remove global-layout bias and express all motion relative to the current vehicle pose, we convert every point into the ego-vehicle coordinate system using the rotation and translation of the current pose:
\[
\tilde{\mathbf{p}}_t = (\mathbf{p}_t - \mathbf{p}_0) \, \mathbf{R}_0^\top ,
\]
where $\mathbf{R}_0$ is the rotation matrix of the current pose.  
The past and future ego trajectories are then uniformly sampled at $2$\,Hz, producing
\[
\text{Past: } \{\tilde{\mathbf{p}}^{\,\text{past}}_i\}_{i=1}^{6}, \qquad
\text{Future: } \{\tilde{\mathbf{p}}^{\,\text{future}}_j\}_{j=1}^{10} .
\]

\noindent\textcolor{planning_gold}{{\footnotesize$\blacksquare$}}~\textbf{\textit{Stage 2: Deriving High-Level Intent Labels.}}
Using the future ego-trajectory, we classify the semantic driving intention along two axes: \emph{speed intent} and \emph{path intent}. 

\noindent\textit{1) Speed Intent.} Velocity is estimated via finite differences,
\[
v_t = \frac{\lVert \tilde{p}_{t+1} - \tilde{p}_t \rVert}{\Delta t},
\]
and intent is determined by thresholding velocity change:
\begin{itemize}
\item \texttt{STOP}: all velocities below a small threshold;
\item \texttt{ACCELERATE}: final velocity exceeds initial velocity significantly;
\item \texttt{DECELERATE}: the reverse case;
\item \texttt{KEEP}: speed remains approximately constant.
\end{itemize}

\noindent\textit{2) Path Intent.} Lateral displacement in the ego frame determines the future path:
\[
\Delta x = x^{\text{future}}_{10}, \qquad
\Delta y = y^{\text{future}}_{10}.
\]
Path direction is determined by comparing the magnitude of lateral drift against a fixed threshold, while ensuring the longitudinal component provides sufficient forward motion stability.
The labeling rule is:
\begin{itemize}
\item \texttt{LEFT\_TURN}: $\Delta x$ exceeds a positive lateral threshold, indicating a clear drift toward the ego-vehicle’s left side;
\item \texttt{RIGHT\_TURN}: $\Delta x$ falls below a negative threshold, indicating motion toward the right side;
\item \texttt{STRAIGHT}: the longitudinal component dominates, i.e., $|\Delta x|$ is small relative to $|\Delta y|$;
\item \texttt{UNKNOWN}: motion is too small or inconsistent to reveal a reliable direction.
\end{itemize}

\noindent\textcolor{planning_gold}{{\footnotesize$\blacksquare$}}~\textbf{\textit{Stage 3: Building Planning QA Samples.}}
Two complementary QA tasks are generated for each frame:
\begin{itemize}
\item \textit{High-level intent planning}:  
The model receives the RGB image, event representation, and current speed, output the two-token driving decision
\begin{center}
\texttt{<SPEED>, <PATH>}
\end{center}
(e.g., ``\texttt{DECELERATE, RIGHT\_TURN}'').  
The ground truth comes directly from the rule-based labels above.

\item \textit{Trajectory forecasting}:  
The model is given the past $3$\,s of ego-motion in ego coordinates and must predict the next $10$ waypoints at $0.5$\,s intervals.  
The ground-truth future trajectory is formatted as a sequence of
\[
[x_j,\, y_j] \quad (j=1,\dots,10),
\]
with four-decimal precision.
\end{itemize}

\noindent\textcolor{planning_gold}{{\footnotesize$\blacksquare$}}~\textbf{\textit{Stage 4: Conversion into Qwen Training Format.}}
Each frame produces two Qwen-style conversation samples.  
The human message describes the driving context, the required output format, and either \textit{(1) the intent-choice rules} or \textit{(2) the past-waypoint history}. The assistant message contains only the strict ground-truth output, without explanations. All samples reference the corresponding RGB image and its aligned event stream, producing a unified, multimodal planning dataset suitable for both training and evaluation. The prompting template for high-level intent planning inference is shown in \cref{fig:infer_planning_intent}, and for trajectory forecasting is shown in \cref{fig:infer_planning_waypoint}.

\section{Complete Experimental Details}
\subsection{Baselines}

\begin{table*}[t]
\centering
\caption{
Per-category comparison in \includegraphics[width=0.022\linewidth]{icons/perception.png}~\textbf{Perception} task on \eventdrive \textbf{DSEC} subset. We report per-category results for \textbf{Scene Type}, \textbf{Visibility}, \textbf{Traffic Flow}, \textbf{Weather}, \textbf{Traffic Light}, and \textbf{Road Condition}, respectively.
}
\resizebox{\linewidth}{!}{
\begin{tabular}{r|cccccc}
\toprule
\textbf{Method} & \textbf{Scene Type} & \textbf{Visibility} & \textbf{Traffic Flow} & \textbf{Weather} & \textbf{Traffic Light} & \textbf{Road Condition}
\\
\midrule\midrule
\rowcolor{event_green!15}\multicolumn{7}{c}{\textcolor{event_green}{\textbf{Event-based Models}}} 
\\
EventGPT-7B \cite{liu2024eventgpt} &  $55.12$ & $52.15$  & $63.26$  & $31.89$  & $47.54$  & $58.42$    \\
\eventdrive-VLM   &   $72.83$ & $61.25$ & $74.46$ & $43.12$ & $59.74$ & $63.25$    \\
\midrule
\rowcolor{frame_red!15}\multicolumn{7}{c}{\textcolor{frame_red}{\textbf{Frame-based Models}}} 
\\
LLaVA-v1.6-Mistral-7B-hf \cite{llavamistral} & $67.63$ & $56.50$ & $70.81$  & $36.30$  &  $52.11$  & $68.57$
\\
LLaVA-OneVision-1.5-8B \cite{llavaonevision} & $90.46$ & $83.35$ & $91.96$  & $38.45$  &  $84.57$  & $96.07$  
\\
InternVL2.5-8B \cite{chen2024expanding} & $87.09$ & $87.56$ & $92.98$  & $35.17$  &  $86.81$  & $80.92$ 
\\
InternVL3-8B \cite{zhu2025internvl3} & $90.18$ & $83.63$ & $84.10$  & $31.71$  &  $88.59$  & $93.45$ 
\\
Qwen2.5-VL-7B-Instruct \cite{qwen2.5-VL} & $86.62$ & $80.17$ & $84.57$  & $40.13$  &  $86.44$   & $88.31$ 
\\
Qwen2.5-VL-7B-Instruct* \cite{qwen2.5-VL} & $93.52$ & $86.73$ & $87.21$ & $36.88$ & $89.87$ & $94.91$
\\
\midrule
\rowcolor{fusion_blue!15}\multicolumn{7}{c}{\textcolor{fusion_blue}{\textbf{Event + Frame Models}}} 
\\
\eventdrive-VLM       & $96.12$ & $89.05$ & $89.93$ & $46.95$ & $94.14$ & $96.42$  \\
\bottomrule
\end{tabular}
}
\label{tab:supp_perception_dsec}
\end{table*}

To comprehensively assess the contributions of event sensing across the driving stack, we compare our model with a broad set of existing vision–language systems, covering both frame-based and event-based paradigms. These baselines allow us to isolate the performance differences arising from sensing modality, temporal fidelity, and instruction tuning strategy under our event–frame benchmark.

\subsubsection{Frame-based Methods} For frame-based evaluation, we select several fully open-source VLMs that represent the current frontier in multimodal reasoning. These models differ substantially in visual encoder capacity, instruction-tuning depth, and the ability to handle multiple images or videos, providing a diverse reference for assessing robustness in event-heavy driving scenes.

\begin{itemize}
    \item \textbf{LLaVA-v1.6-Mistral-7B} \cite{llavamistral} is a multimodal instruction-following model that aligns a Mistral-7B \cite{jiang2023mistral7b} backbone with a large collection of curated image–text and GPT-generated interaction data. The model integrates a lightweight visual encoder with an autoregressive language model, enabling competitive zero-shot reasoning and general-purpose VQA capabilities. Although trained on diverse imagery and conversational tasks, its perception is primarily frame-centric, offering a useful reference for evaluating robustness in event-heavy or motion-degraded driving scenes.

    \item \textbf{LLaVA-OneVision-1.5} \cite{llavaonevision} is an open multimodal model family trained on large-scale, high-quality image–text corpora with native-resolution images, improving recognition of fine-grained visual details. Its training framework provides efficient scaling through Megatron-LM \cite{shoeybi2020megatron} and long-sequence optimization, enabling strong zero-shot performance across diverse multimodal benchmarks. As a frame-centric LMM with broad instruction-tuning coverage, it offers a representative baseline for assessing general-purpose visual reasoning under driving scenarios.

    \item \textbf{InternVL 2.5} \cite{chen2024expanding} is a multimodal model family built upon a ViT–MLP–LLM architecture, pairing an incrementally trained InternViT \cite{gao2024mini, chen2024far, chen2024internvl} encoder with modern language backbones such as InternLM 2.5 \cite{cai2024internlm2} and Qwen2.5 \cite{qwen2025qwen25}. The model adopts dynamic high-resolution tiling and pixel-unshuffle token reduction, enabling efficient processing of single images, multi-image inputs, and videos. Its training pipeline combines cross-modal warm-up, optional vision-encoder refinement, and large-scale instruction tuning with strict data filtering, yielding strong general-purpose visual and multimodal reasoning. As a frame-based LMM with high-resolution processing, InternVL provides a competitive benchmark for evaluating robustness in event-rich driving scenarios.

    \item \textbf{InternVL 3} \cite{zhu2025internvl3} extends the ViT–MLP–LLM architecture of InternVL 2.5 with native multimodal pre-training, allowing the vision and language components to learn jointly from interleaved image–text and video–text corpora. It incorporates an incrementally trained InternViT \cite{gao2024mini, chen2024far, chen2024internvl} encoder and introduces variable visual position encoding for improved long-context reasoning across multi-image and video inputs. Enhanced supervised fine-tuning and mixed preference optimization further strengthen its multimodal understanding and CoT performance. As a high-capacity frame-based baseline, InternVL3 offers a strong reference for evaluating perception and reasoning under driving scenarios.

    \item \textbf{Qwen2.5-VL} \cite{qwen2.5-VL} is an upgraded vision–language model that enhances its ViT encoder and temporal modeling to support dynamic-resolution images and variable-rate video inputs. The model demonstrates strong capabilities in structured visual understanding, including text-rich imagery, charts, and layout reasoning, and supports fine-grained localization through bounding boxes and keypoints. Its instruction-tuned design further enables tool-driven visual reasoning and long-video event analysis. As a frame-centric VLM, Qwen2.5-VL provides a strong baseline for assessing semantic and spatial reasoning in driving scenarios.
    
\end{itemize}

\begin{table*}[t]
\centering
\caption{
Per-category comparison in \includegraphics[width=0.022\linewidth]{icons/perception.png}~\textbf{Perception} task on \eventdrive \textbf{M3ED} subset. We report per-category results for \textbf{Scene Type}, \textbf{Visibility}, \textbf{Traffic Flow}, \textbf{Weather}, \textbf{Traffic Light}, and \textbf{Road Condition}, respectively.
}
\resizebox{\linewidth}{!}{
\begin{tabular}{r|cccccc}
\toprule
\textbf{Method} & \textbf{Scene Type} & \textbf{Visibility} & \textbf{Traffic Flow} & \textbf{Weather} & \textbf{Traffic Light} & \textbf{Road Condition}
\\
\midrule\midrule
\rowcolor{event_green!15}\multicolumn{7}{c}{\textcolor{event_green}{\textbf{Event-based Models}}} 
\\
EventGPT-7B \cite{liu2024eventgpt} & $70.76$ & $38.66$ & $55.58$ & $57.38$ & $50.01$ & $57.43$  \\
\eventdrive-VLM          & $88.54$ & $47.18$ & $67.81$ & $70.01$ & $56.61$ & $72.27$   \\
\midrule
\rowcolor{frame_red!15}\multicolumn{7}{c}{\textcolor{frame_red}{\textbf{Frame-based Models}}} 
\\
LLaVA-v1.6-Mistral-7B-hf \cite{llavamistral} & $87.04$ & $41.67$ & $59.90$  & $61.84$  &  $32.34$  & $72.67$
\\
LLaVA-OneVision-1.5-8B \cite{llavaonevision} & $98.62$ & $77.78$ & $88.09$  & $82.34$  &  $74.09$  & $97.20$  
\\
InternVL2.5-8B \cite{chen2024expanding} & $99.59$ & $72.85$ & $86.01$  & $82.45$  &  $75.73$  & $83.76$ 
\\
InternVL3-8B \cite{zhu2025internvl3} & $99.29$ & $78.90$ & $78.08$  & $86.15$  &  $77.07$  & $98.13$ 
\\
Qwen2.5-VL-7B-Instruct \cite{qwen2.5-VL} & $98.13$ & $53.47$ & $81.18$ & $64.13$   &  $79.57$ & $85.14$ 
\\
Qwen2.5-VL-7B-Instruct* \cite{qwen2.5-VL} & $98.02$ & $68.86$ & $89.36$ & $70.59$ & $87.59$ & $93.72$ 
\\
\midrule
\rowcolor{fusion_blue!15}\multicolumn{7}{c}{\textcolor{fusion_blue}{\textbf{Event + Frame Models}}}
\\
\eventdrive-VLM       & $98.28$ & $72.45$ & $91.42$ & $72.22$ & $89.61$ & $95.88$   \\
\bottomrule
\end{tabular}
}
\label{tab:supp_perception_m3ed}
\end{table*}

All frame-based models are evaluated in a zero-shot manner to measure out-of-the-box generalization under low light, motion blur, and rapid agent dynamics. In addition, to explicitly quantify the gain brought by event signals, we fine-tune a Qwen2.5-VL-7B model using the same instruction-tuning protocol as our method, serving as a frame-only counterpart that shares the backbone of our \eventdrive-VLM.

\subsubsection{Event-based Methods}
Event-based multimodal models remain scarce due to the absence of publicly released training frameworks. Systems such as EventVL~\cite{li2025eventvl} provide neither checkpoints nor code, leaving EventGPT~\cite{liu2024eventgpt} as the only available event-driven LLM for direct comparison.

EventGPT is designed to extend language-model reasoning to asynchronous event streams through a three-stage training paradigm involving image–text warm-up, large-scale synthetic event–text pre-training, and instruction tuning on the Event-Chat dataset. Its data construction relies heavily on synthetic corpora such as N-ImageNet-Chat and N-ImageNet-Instruction, with only a smaller portion drawn from real event recordings. By aligning sparse spatio-temporal event representations with language, EventGPT demonstrates improved scene summarization and reasoning under low-light and high-motion conditions where RGB-based models deteriorate.

Because EventGPT’s training pipeline, evaluation split, and testing code are not publicly released, we can only evaluate the model in a zero-shot manner on the subset of its dataset provided in the official repository, and the resulting numbers should therefore be interpreted with caution. As an event-only model optimized for temporal cues rather than appearance semantics, its performance serves as a qualitative reference rather than a fully standardized comparison within our benchmark.

\subsubsection{Event-Frame Fusion Methods}
For event–frame fusion approaches, existing multimodal event LLMs such as LLaFEA \cite{zhou2025llafea} have not released code or checkpoints, preventing controlled reproduction or fine-tuning. Thus, our \eventdrive-VLM serves as the primary event–frame fusion baseline, enabling a structured investigation of how asynchronous event cues complement RGB perception across perception, understanding, prediction, and planning.

\begin{table*}[t]
\centering
\caption{
Per-category comparison in \includegraphics[width=0.022\linewidth]{icons/perception.png}~\textbf{Perception} task on \eventdrive \textbf{PKU} subset. We report per-category results for \textbf{Scene Type}, \textbf{Visibility}, \textbf{Traffic Flow}, \textbf{Weather}, \textbf{Traffic Light}, and \textbf{Road Condition}, respectively.
}
\resizebox{\linewidth}{!}{
\begin{tabular}{r|cccccc}
\toprule
\textbf{Method} & \textbf{Scene Type} & \textbf{Visibility} & \textbf{Traffic Flow} & \textbf{Weather} & \textbf{Traffic Light} & \textbf{Road Condition}
\\
\midrule\midrule
\rowcolor{event_green!15}\multicolumn{7}{c}{\textcolor{event_green}{\textbf{Event-based Models}}} 
\\
EventGPT-7B \cite{liu2024eventgpt} & $79.67$ & $36.48$ & $49.11$ & $41.32$ & $32.41$ & $74.50$   \\
\eventdrive-VLM       & $85.13$ & $48.56$ & $58.64$ & $54.34$ & $38.70$ & $88.96$   \\
\midrule
\rowcolor{frame_red!15}\multicolumn{7}{c}{\textcolor{frame_red}{\textbf{Frame-based Models}}} 
\\
LLaVA-v1.6-Mistral-7B-hf \cite{llavamistral} & $76.32$ & $34.95$ & $47.05$  & $39.58$  &  $31.05$  & $71.37$
\\
LLaVA-OneVision-1.5-8B \cite{llavaonevision} & $84.84$ & $52.53$ & $78.95$  & $62.21$  &  $67.47$  & $88.53$  
\\
InternVL2.5-8B \cite{chen2024expanding} & $93.47$ & $49.47$ & $76.11$  & $74.21$  &  $72.84$  & $89.05$ 
\\
InternVL3-8B \cite{zhu2025internvl3} & $93.26$ & $45.05$ & $71.89$  & $69.05$  &  $74.04$  & $92.95$ 
\\
Qwen2.5-VL-7B-Instruct \cite{qwen2.5-VL} & $80.63$ & $54.53$ & $61.68$ & $41.58$ & $73.47$ & $62.53$
\\
Qwen2.5-VL-7B-Instruct* \cite{qwen2.5-VL} & $93.04$ & $56.31$ & $75.00$ & $65.56$ & $84.34$ & $81.03$ 
\\
\midrule
\rowcolor{fusion_blue!15}\multicolumn{7}{c}{\textcolor{fusion_blue}{\textbf{Event + Frame Models}}}
\\
\eventdrive-VLM        & $93.73$ & $61.54$ & $77.98$ & $68.16$ & $87.69$ & $84.24$   \\
\bottomrule
\end{tabular}
}
\label{tab:supp_perception_pku}
\end{table*}

\subsection{Evaluation Protocol}
We adopt a unified evaluation protocol across the four task families in \eventdrive. Below, we provide the complete metric definitions and computation procedures.

\subsubsection{\perception Evaluation} Perception evaluates scene-level attributes using multiple-choice questions.
For evaluation, each original question is decomposed into two split prompts: one requests the option letter and the other requests the textual label. The two predictions are paired using their shared original-question identifier.
Each paired answer consists of two components:
a choice identifier $c \in \{A, B, C, D\}$ and a textual label $t$ describing the attribute class (e.g., ``Urban'', ``Tunnel'', ``Night''). 
Given a ground-truth answer $y = (c, t)$ and a model prediction $\hat{y} = (\hat{c}, \hat{t})$, 
we first normalize both components by removing prefixes such as ``Answer:'', stripping punctuation, 
collapsing repeated whitespace, and parsing the prediction into its letter and text components. 
Choice identifiers are converted to uppercase, and textual labels are compared case-insensitively.

A prediction is considered correct only when both the choice identifier and textual label match:
\begin{equation}
\mathrm{Correct}(i) = 
\mathbf{1}\left(c_i = \hat{c}_i \;\land\; t_i = \hat{t}_i \right),
\end{equation}
where $\mathbf{1}( \cdot )$ is the indicator function, and the overall perception accuracy is:
\begin{equation}
\texttt{Acc}_{\mathrm{perc}} = 
\frac{1}{N} \sum_{i=1}^{N} \mathrm{Correct}(i) .
\end{equation}
This strict matching prevents degenerate strategies such as predicting only the textual label without choosing the correct option.
 
Each perception question belongs to a predefined scene category (e.g., scene type, traffic flow, weather). 
For each category $k$, accuracy is computed as:
\begin{equation}
\texttt{Acc}(k) =
\frac{\sum_{i \in k} \mathrm{Correct}(i)}{|k|}.
\end{equation}
Category-wise results allow us to analyze robustness under challenging conditions, such as tunnel transitions, nighttime scenes, or strong motion blur.

\subsubsection{\understanding Evaluation} Understanding includes two complementary components: 
\textit{1) object-level multiple-choice question answering}, and 
\textit{2) spatial grounding via bounding box localization}. 
Each sample contains either a categorical QA label or a ground-truth bounding box, depending on the task type.

As in perception, each non-grounding question is decomposed into an option-letter prompt and a textual-label prompt. The two predictions are paired using their shared original-question identifier.
Each paired answer consists of a choice identifier $c \in \{A, B, C, D\}$ and a textual label $t$.
Given a ground-truth answer $y = (c, t)$ and a normalized prediction 
$\hat{y} = (\hat{c}, \hat{t})$, 
we evaluate correctness using a strict letter match and a soft textual match.
The QA accuracy is:

\begin{equation}
\texttt{Acc}_{\mathrm{under}} = 
\frac{1}{N}
\sum_{i=1}^{N}
\mathbf{1}\!\left( 
c_i = \hat{c}_i
\;\land\;
\texttt{SoftMatch}(t_i, \hat{t}_i)
\right),
\end{equation}
where $\texttt{SoftMatch}(\cdot)$ is a relaxed matching function that:
(1) removes articles (``a'', ``an'', ``the''),  
(2) strips punctuation,  
(3) normalizes whitespace and commas,  
(4) computes a Jaccard similarity over token sets.  
A pair $(t_i, \hat{t}_i)$ is considered matched when

\begin{equation}
\mathrm{Jaccard}(t_i, \hat{t}_i) > 0.8.
\end{equation}
This rule mitigates minor formatting variations frequently produced by language models.

For \textit{grounding} samples, the model predicts a bounding box 
$\hat{b} = (\hat{x}, \hat{y}, \hat{w}, \hat{h})$,
while the ground truth is 
$b = (x, y, w, h)$.
We compute Intersection-over-Union (IoU) as:

\begin{equation}
\mathrm{IoU}(b, \hat{b}) = 
\frac{|b \cap \hat{b}|}{|b \cup \hat{b}|}.
\end{equation}

Two grounding metrics are reported, including \textit{Top-1 Localization Accuracy}:
\begin{equation}
\texttt{Acc}_{\mathrm{gd}} =
\frac{1}{N_{\mathrm{gd}}}
\sum_{i=1}^{N_{\mathrm{gd}}}
\mathbf{1}\!\left(
\mathrm{IoU}(b_i, \hat{b}_i) \geq \tau
\right),
\quad \tau = 0.6,
\end{equation}
and \textit{Mean IoU}:
\begin{equation}
\mathrm{mIoU} = 
\frac{1}{N_{\mathrm{gd}}}
\sum_{i=1}^{N_{\mathrm{gd}}}
\mathrm{IoU}(b_i, \hat{b}_i).
\end{equation}

Each understanding sample also belongs to a semantic category 
(e.g., status, ego-relation, grounding).  
For a category $k$, accuracy is computed as:
\begin{equation}
\texttt{Acc}(k) 
= 
\frac{\sum_{i \in k} 
 \text{Correct}(i) }
{|k|}.
\end{equation}
This provides fine-grained insight into how event signals affect semantic understanding and spatial grounding across different object-level reasoning tasks.

\begin{table*}[t]
\centering
\caption{
\textbf{Comparison across four driving–reasoning tasks} in the \eventdrive \textbf{Hard} benchmark. For \textbf{perception}, we report \texttt{QA Accuracy} on three subsets. For \textbf{understanding}, we report \texttt{QA Accuracy}, grounding Top-1 \texttt{Accuracy} at IoU $0.6$, and \textbf{mIoU}. For \textbf{prediction} and \textbf{planning}, we report \texttt{Speed Accuracy} and \texttt{Path Accuracy}, and for planning, we also report the mean \textbf{L2 Error} (lower is better). All scores are reported in percentage (\%), except for L2 Error in meters. Best results are shown in \textbf{bold} and 2nd-best results are \underline{underlined}.
}
\vspace{-3mm}
\resizebox{\linewidth}{!}{
\begin{tabular}{r|ccc|ccc|cc|ccc}
\toprule
\multirow{2}{*}{\textbf{Method}} & 
\multicolumn{3}{c|}{\includegraphics[width=0.022\linewidth]{icons/perception.png}~\textbf{Perception}} & 
\multicolumn{3}{c|}{\includegraphics[width=0.022\linewidth]{icons/understanding.png}~\textbf{Understanding}}  & 
\multicolumn{2}{c|}{\includegraphics[width=0.022\linewidth]{icons/prediction.png}~\textbf{Prediction}}  & 
\multicolumn{3}{c}{\includegraphics[width=0.022\linewidth]{icons/planning.png}~\textbf{Planning}}  \\
 & \texttt{Acc@D} & \texttt{Acc@M} & \texttt{Acc@P}
 & \texttt{Acc} & \texttt{Acc@60} & \textbf{mIoU} 
 & \texttt{Speed} & \texttt{Path} 
 & \texttt{Speed} & \texttt{Path} & \textbf{L2 Error} \\
\midrule\midrule
\rowcolor{event_green!15}\multicolumn{12}{c}{\textcolor{event_green}{\textbf{Event-based Models}}} 
\\
EventGPT-7B \cite{liu2024eventgpt} & $49.09$  & $55.00$  & $52.83$  & $36.96$  & $2.89$  & $3.67$  & $14.26$  & $59.24$  &  $24.82$ & $69.64$  & $12.24 $ \\
\eventdrive-VLM                  & $67.30$  & $66.22$  & $66.93$  & $52.17$  & $34.22$  & $29.50$  &  \underline{$28.56$} & $76.48$  &  $39.23$ & $85.53$  &  $7.26$ \\
\midrule
\rowcolor{frame_red!15}\multicolumn{12}{c}{\textcolor{frame_red}{\textbf{Frame-based Models}}} 
\\
LLaVA-v1.6-Mistral-7B-hf \cite{llavamistral} & $55.64$ & $61.55$ & $50.45$  & $36.69$  & $15.23$  & $25.3$  & $20.10$  & $23.25$  & $10.43$  & \underline{$88.59$} & $7.31$  
\\
LLaVA-OneVision-1.5-8B \cite{llavaonevision} & \underline{$80.00$} & $85.36$ & $76.26$  &  $57.09$ & $2.65$  & $16.1$  &  $15.23$ & $55.84$  & \underline{$46.44$}
& $41.12$  & $8.31$  
\\
InternVL2.5-8B \cite{chen2024expanding} & $75.95$ & \underline{$86.27$} & $\mathbf{80.15}$  & $58.28$  & $0.18$  & $2.21$  &  $27.05$ & $78.52$  &  $34.02$ &  $51.75$ &  $11.08$ 
\\
InternVL3-8B \cite{zhu2025internvl3} & $74.11$ & $84.09$ & $75.61$ & \underline{$58.68$}  & $0.16$ & $1.51$  &  $13.71$ & \underline{$78.58$}  & $33.78$ &  $83.62$ & $8.97$
\\
Qwen2.5-VL-7B-Instruct \cite{qwen2.5-VL} & $72.25$ & $77.14$ & $60.76$ & $43.05$ & $27.15$ & $35.12$ & $11.68$ & $70.15$ & $43.23$ & $82.62$ & $7.91$ 
\\
Qwen2.5-VL-7B-Instruct* \cite{qwen2.5-VL} &  $77.94$ & $83.52$  & $68.88$  &  $50.77$ & \underline{$38.15$}  & \underline{$40.89$}  & $20.76$  &  $75.46$ & $46.02$  & $86.44$  & \underline{$6.02$}  
\\
\midrule
\rowcolor{fusion_blue!15}\multicolumn{12}{c}{\textcolor{fusion_blue}{\textbf{Event + Frame Models}}} 
\\
\eventdrive-VLM       & $\mathbf{82.43}$  & $\mathbf{86.09}$  & \underline{$76.94$}  &  $\mathbf{59.04}$ &  $\mathbf{45.86}$ & $\mathbf{49.66}$  & $\mathbf{29.48}$  & $\mathbf{79.32}$  & $\mathbf{52.45}$  & $\mathbf{88.64}$ &  $\mathbf{5.41}$ \\
\bottomrule
\end{tabular}
}
\label{tab:eventdrive_hard}
\end{table*}

\subsubsection{\prediction Evaluation} 
Prediction assesses the model’s ability to infer short-term motion tendencies of dynamic agents. Each original sample contains two categorical labels: a \emph{speed intent} $y^{\mathrm{spd}}$ and a \emph{path intent} $y^{\mathrm{path}}$. During inference, the two intents are queried separately and paired using their shared original-sample identifier.
Given the paired prediction $(\hat{y}^{\mathrm{spd}}, \hat{y}^{\mathrm{path}})$, we compute intent accuracy for each component as well as their joint correctness.

Since language models may output free-form text such as
``A ACCELERATE, C STRAIGHT'' or ``the vehicle will keep straight,''
we normalize predictions by:
(1) removing punctuation and repeated whitespace,
(2) uppercasing tokens, and 
(3) extracting the canonical intent words using a predefined vocabulary for speed and path.
This ensures robustness to minor formatting variations.

A speed prediction is correct when
$\hat{y}^{\mathrm{spd}} = y^{\mathrm{spd}}$.
Thus,
\begin{equation}
\texttt{Acc}_{\mathrm{spd}} =
\frac{1}{N}
\sum_{i=1}^{N}
\mathbf{1}\!\left(
\hat{y}^{\mathrm{spd}}_i = y^{\mathrm{spd}}_i
\right).
\end{equation}

Similarly, path correctness is evaluated as:
\begin{equation}
\texttt{Acc}_{\mathrm{path}} =
\frac{1}{N}
\sum_{i=1}^{N}
\mathbf{1}\!\left(
\hat{y}^{\mathrm{path}}_i = y^{\mathrm{path}}_i
\right).
\end{equation}

\begin{table*}[t]
\centering
\caption{
Per-category comparison in the \includegraphics[width=0.022\linewidth]{icons/understanding.png}~\textbf{Understanding} task on the \eventdrive benchmark. We report per-category results for \textbf{Object Awareness}, \textbf{Appearance}, \textbf{Status}, \textbf{Relation-to-Viewer}, and \textbf{Relation-to-Others}, respectively.
}
\resizebox{\linewidth}{!}{
\begin{tabular}{r|ccccc}
\toprule
\textbf{Method} & \textbf{Object Awareness} & \textbf{Appearance} & \textbf{Status} & \textbf{Relation-to-Viewer} & \textbf{Relation-to-Others}
\\
\midrule\midrule
\rowcolor{event_green!15}\multicolumn{6}{c}{\textcolor{event_green}{\textbf{Event-based Models}}} 
\\
EventGPT-7B \cite{liu2024eventgpt} & $41.75$ & $46.03$ & $41.55$ & $34.90$ & $29.69$    \\
\eventdrive-VLM          & $58.73$ & $64.95$ & $57.98$ & $47.47$ & $41.92$    \\
\midrule
\rowcolor{frame_red!15}\multicolumn{6}{c}{\textcolor{frame_red}{\textbf{Frame-based Models}}} 
\\
LLaVA-v1.6-Mistral-7B-hf \cite{llavamistral} & $31.20$ & $64.31$ & $45.34$ & $27.83$ & $33.15$
\\
LLaVA-OneVision-1.5-8B \cite{llavaonevision} & $66.76$ & $73.84$ & $65.93$ & $53.95$ & $47.63$
\\
InternVL2.5-8B \cite{chen2024expanding} & $66.39$ & $84.82$ & $62.23$ & $43.76$ & $42.89$
\\
InternVL3-8B \cite{zhu2025internvl3}   & $66.00$ & $85.27$ & $67.09$ & $44.96$ & $39.74$
\\
Qwen2.5-VL-7B-Instruct \cite{qwen2.5-VL} & $58.99$ & $57.57$ & $55.74$ & $37.15$ & $40.47$
\\
Qwen2.5-VL-7B-Instruct* \cite{qwen2.5-VL} & $68.97$ & $67.31$ & $65.17$ & $43.43$ & $47.32$    
\\
\midrule
\rowcolor{fusion_blue!15}\multicolumn{6}{c}{\textcolor{fusion_blue}{\textbf{Event + Frame Models}}}
\\
\eventdrive-VLM      & $77.30$ & $75.39$ & $73.01$ & $48.68$ & $53.02$    \\
\bottomrule
\end{tabular}
}
\label{tab:supp_understanding}
\end{table*}

\begin{table*}[t]
\centering
\caption{
Per-category comparison in the \includegraphics[width=0.022\linewidth]{icons/understanding.png}~\textbf{Understanding} task on the \eventdrive \textbf{hard} subset. We report per-category results for \textbf{Object Awareness}, \textbf{Appearance}, \textbf{Status}, \textbf{Relation-to-Viewer}, and \textbf{Relation-to-Others}, respectively.
}
\resizebox{\linewidth}{!}{
\begin{tabular}{r|ccccc}
\toprule
\textbf{Method} & \textbf{Object Awareness} & \textbf{Appearance} & \textbf{Status} & \textbf{Relation-to-Viewer} & \textbf{Relation-to-Others}
\\
\midrule\midrule
\rowcolor{event_green!15}\multicolumn{6}{c}{\textcolor{event_green}{\textbf{Event-based Models}}} 
\\
EventGPT-7B \cite{liu2024eventgpt} & $36.47$ & $44.22$ & $50.61$ & $24.64$ & $28.85$    \\
\eventdrive-VLM  & $58.36$ & $66.48$ & $71.44$ & $45.47$ & $19.10$   \\
\midrule
\rowcolor{frame_red!15}\multicolumn{6}{c}{\textcolor{frame_red}{\textbf{Frame-based Models}}} 
\\
LLaVA-v1.6-Mistral-7B-hf \cite{llavamistral} & $35.10$ & $43.05$ & $49.01$ & $23.84$ & $32.45$
\\
LLaVA-OneVision-1.5-8B \cite{llavaonevision} & $66.23$ & $50.33$ & $66.23$ & $54.97$ & $47.68$
\\
InternVL2.5-8B \cite{chen2024expanding} & $66.23$ & $64.24$ & $76.82$ & $46.36$ & $37.75$
\\
InternVL3-8B \cite{zhu2025internvl3}   & $69.44$ & $63.87$ & $78.55$ & $45.06$ & $36.48$ 
\\
Qwen2.5-VL-7B-Instruct \cite{qwen2.5-VL} & $60.93$ & $39.07$ & $57.62$ & $21.85$ & $35.76$
\\
Qwen2.5-VL-7B-Instruct* \cite{qwen2.5-VL} & $69.56$ & $47.31$ & $64.42$ & $25.96$ & $46.59$  
\\
\midrule
\rowcolor{fusion_blue!15}\multicolumn{6}{c}{\textcolor{fusion_blue}{\textbf{Event + Frame Models}}}
\\
\eventdrive-VLM  & $78.67$ & $59.24$ & $71.04$ & $31.98$ & $54.29$     \\
\bottomrule
\end{tabular}
}
\label{tab:supp_understanding_hard_split}
\end{table*}

A prediction is considered fully correct only when both the speed and path intents match:
\begin{equation}
\texttt{Acc}_{\mathrm{joint}} =
\frac{1}{N}
\sum_{i=1}^{N}
\mathbf{1}\!\left(
\hat{y}^{\mathrm{spd}}_i = y^{\mathrm{spd}}_i
\;\land\;
\hat{y}^{\mathrm{path}}_i = y^{\mathrm{path}}_i
\right).
\end{equation}
This metric reflects the model’s ability to jointly reason about the agent’s longitudinal tendency (speed changes) and lateral evolution (path direction).

\subsubsection{\planning Evaluation} 
Planning evaluates both high-level driving intent and low-level future trajectory prediction. 
Each sample belongs to one of two categories: 
\textit{1) Planning-HighLevel}, which requires predicting ego speed intent and path intent, or 
\textit{2) Planning-Trajectory}, which requires forecasting future ego waypoints over a $5$-second horizon.

For each high-level sample, speed and path intents are queried separately and paired using their shared original-sample identifier. The resulting prediction is
$(\hat{y}^{\mathrm{spd}}, \hat{y}^{\mathrm{path}})$. 
Because language models may produce free-form text (e.g., ``the vehicle will accelerate and turn right''), we normalize predictions by stripping punctuation, collapsing whitespace, uppercasing tokens, 
and extracting intent words from the predefined vocabularies.

Speed and path intent accuracies are computed as:
\begin{equation}
\texttt{Acc}_{\mathrm{spd}} =
\frac{1}{N_{\mathrm{high}}}
\sum_{i=1}^{N_{\mathrm{high}}}
\mathbf{1}\!\left(
\hat{y}^{\mathrm{spd}}_i = y^{\mathrm{spd}}_i
\right),
\end{equation}
\begin{equation}
\texttt{Acc}_{\mathrm{path}} =
\frac{1}{N_{\mathrm{high}}}
\sum_{i=1}^{N_{\mathrm{high}}}
\mathbf{1}\!\left(
\hat{y}^{\mathrm{path}}_i = y^{\mathrm{path}}_i
\right).
\end{equation}
A prediction is fully correct when both components match:
\begin{equation}
\texttt{Acc}_{\mathrm{joint}} =
\frac{1}{N_{\mathrm{high}}}
\sum_{i=1}^{N_{\mathrm{high}}}
\mathbf{1}\!\left(
\hat{y}^{\mathrm{spd}}_i = y^{\mathrm{spd}}_i
\;\land\;
\hat{y}^{\mathrm{path}}_i = y^{\mathrm{path}}_i
\right).
\end{equation}

Each trajectory sample provides 
a sequence of $T=10$ future waypoints at 0.5-second intervals:
\[
W_i = \{(x_{i,t}, y_{i,t})\}_{t=1}^{T},
\qquad
\hat{W}_i = \{(\hat{x}_{i,t}, \hat{y}_{i,t})\}_{t=1}^{T}.
\]
We compute the per-step Euclidean error:
\begin{equation}
e_{i,t} =
\left\lVert 
\begin{pmatrix} x_{i,t} \\ y_{i,t} \end{pmatrix}
-
\begin{pmatrix} \hat{x}_{i,t} \\ \hat{y}_{i,t} \end{pmatrix}
\right\rVert_2 .
\end{equation}
The benchmark reports waypoint errors at 1s, 3s, and 5s, corresponding to indices 
$t=\{2, 6, 10\}$ under a sampling interval $\Delta t = 0.5$ seconds:
\begin{equation}
\texttt{L2}(t) = e_{i,t}.
\end{equation}
The mean trajectory error is:
\begin{equation}
\texttt{L2}_{\mathrm{mean}} =
\frac{1}{T}
\sum_{t=1}^{T} e_{i,t}.
\end{equation}
We report the average of these errors over all trajectory samples.

\begin{table*}[t]
\centering
\caption{
Per-category comparison in \includegraphics[width=0.022\linewidth]{icons/perception.png}~\textbf{Perception} task on \eventdrive \textbf{DSEC} \textbf{hard} subset. We report per-category results for \textbf{Scene Type}, \textbf{Visibility}, \textbf{Traffic Flow}, \textbf{Weather}, \textbf{Traffic Light}, and \textbf{Road Condition}, respectively.
}
\resizebox{\linewidth}{!}{
\begin{tabular}{r|cccccc}
\toprule
\textbf{Method} & \textbf{Scene Type} & \textbf{Visibility} & \textbf{Traffic Flow} & \textbf{Weather} & \textbf{Traffic Light} & \textbf{Road Condition}
\\
\midrule\midrule
\rowcolor{event_green!15}\multicolumn{7}{c}{\textcolor{event_green}{\textbf{Event-based Models}}} 
\\
EventGPT-7B \cite{liu2024eventgpt} & $55.70$ & $59.47$ & $57.59$ & $16.05$ & $50.03$ & $55.70$   \\
EventDrive-VLM                 & $76.36$ & $81.53$ & $78.95$ & $22.00$ & $68.59$ & $76.36$   \\
\midrule
\rowcolor{frame_red!15}\multicolumn{7}{c}{\textcolor{frame_red}{\textbf{Frame-based Models}}} 
\\
LLaVA-v1.6-Mistral-7B-hf \cite{llavamistral} & $69.23$ & $86.15$ & $56.92$ & $7.69$ & $50.77$ & $63.08$
\\
LLaVA-OneVision-1.5-8B \cite{llavaonevision} & $90.77$ & $96.92$ & $93.85$ & $26.15$ & $81.54$ & $90.77$
\\
InternVL2.5-8B \cite{chen2024expanding} & $86.15$ & $98.78$ & $95.38$ & $4.62$ & $90.77$ & $80.00$
\\
InternVL3-8B \cite{zhu2025internvl3} & $92.31$ & $87.69$ & $84.62$ & $4.62$ & $86.15$ & $89.23$
\\
Qwen2.5-VL-7B-Instruct \cite{qwen2.5-VL} & $86.15$ & $98.14$ & $75.38$ & $7.69$ & $86.15$ & $80.00$
\\
Qwen2.5-VL-7B-Instruct* \cite{qwen2.5-VL} & $90.94$ & $98.87$ & $83.31$ & $18.29$ & $89.94$ & $86.30$  
\\
\midrule
\rowcolor{fusion_blue!15}\multicolumn{7}{c}{\textcolor{fusion_blue}{\textbf{Event + Frame Models}}} 
\\
EventDrive-VLM       & $95.29$ & $98.56$ & $85.99$ & $28.17$ & $95.29$ & $91.27$   \\
\bottomrule
\end{tabular}
}
\label{tab:supp_dsec_hard_split}
\end{table*}

\subsection{Additional Implementation Details}

We fine-tune \eventdrive-VLM on top of Qwen2.5-VL-7B Instruct, augmented with a pretrained MaxViT–RNN \cite{gehrig2023rvt} event backbone. The event encoder adopts a multi-horizon voxelization scheme with three temporal bin sizes $\mathcal{B}=\{20, 50, 100\}$ to accommodate varying event sparsity and motion dynamics across datasets. Each voxel grid is processed by a stacked MaxViT-RNN encoder, whose output feature
maps are aggregated through a lightweight Event Q-Former, and a single linear layer mapping the $512$-d event features into the LLM hidden space ($2048$-d). The RGB vision tower of Qwen2.5-VL is kept frozen throughout all experiments.

\noindent\textbf{Training Strategy.}
Training is performed using the \texttt{transformer} framework with several
infrastructure optimizations. We adopt AdamW \cite{kingma2015adam} with a cosine learning schedule and a warmup ratio of $0.03$.
The learning rate is decoupled across modules: the event encoder and event Q-former use $1\times10^{-4}$, and the LLM layers use a conservative $2\times10^{-7}$. All experiments use \textbf{bf16}-precision and enable gradient checkpointing to reduce memory cost.

\noindent\textbf{Packed Multimodal Training.}
To support long multimodal sequences at $4096$ tokens, we employ a packed-sequence data
pipeline. All modalities (RGB frames, events) are flattened into a unified token
stream following Qwen2.5-VL’s format. During batching, visual tokens are concatenated across samples and re-located by per-sample \texttt{position\_ids}, computed using an extended 2D RoPE scheme that also supports events.

\noindent\textbf{Fused FlashAttention.}
We replace the default attention kernels in Qwen2.5-VL with
FlashAttention~2 \cite{dao2023flashattention} for both text-only and multimodal tokens. Because event patches greatly increase the local sequence length,
we adopt the variable-length attention API and
modify the causal mask update in Qwen2.5-VL to support packed multimodal inputs.
This reduces attention overhead by over 40\% and enables training at the full 4096-token context.

\noindent\textbf{Token-Level Training Masking.}
To prevent the LLM from predicting pixels or events, all labels at positions
corresponding to \texttt{<|image\_pad|>}, \texttt{<|event|>}, and \texttt{<|video\_pad|>} tokens are replaced by
$\texttt{IGNORE\_INDEX}$.  
Loss is computed only on natural language tokens:
\[
\mathcal{L} = 
\mathrm{CE}\!\left(
\mathrm{shift}( \mathbf{y}^{\text{logits}} ),
\mathrm{shift}( \mathbf{y}^{\text{target}} )
\right).
\]

\noindent\textbf{Inference.}
At test time, only the first forward pass processes visual tokens.
Subsequent decoding steps remove all image/event features to reduce overhead.
Inference uses the same 2D-RoPE indexing code as training, ensuring exact consistency.

Overall, these design choices allow \eventdrive-VLM to fuse asynchronous event streams with RGB context at scale, while maintaining full compatibility with Qwen2.5-VL’s instruction-following behavior.

\subsection{Additional Quantitative Results}

\noindent \textbf{\perception Per-category Results.} Across all three datasets, the per-category comparisons in \cref{tab:supp_perception_dsec,tab:supp_perception_m3ed,tab:supp_perception_pku} reveal several consistent patterns.
First, pure event-based models (EventGPT-7B) struggle with appearance-heavy categories such as Scene Type, Traffic Light, and Road Condition, where semantic cues depend substantially on spatial textures and color information, confirming that event streams alone are insufficient for fine-grained semantic reasoning.

Frame-based VLMs demonstrate strong improvements in most categories, particularly Scene Type, Traffic Flow, and Road Condition, where global illumination and texture structure are crucial. Large models such as InternVL3-8B and Qwen2.5-VL-7B-Instruct consistently exceed $80\%$ on these categories across datasets. However, they degrade notably on event-favored categories, highlighting their weakness under motion blur and extreme lighting.

Our \eventdrive-VLM shows the strongest and most balanced performance across all six categories. The Event + Frame fusion consistently outperforms both single-modality baselines, with substantial gains on Visibility and Weather. At the same time, performance on appearance-dominant categories (e.g., Road Condition, Traffic Light) matches or exceeds the best frame-based VLMs. These results confirm that event–frame fusion enables robust scene recognition across both texture-dependent and motion/illumination-sensitive conditions.

\begin{table*}[t]
\centering
\caption{
Per-category comparison in \includegraphics[width=0.022\linewidth]{icons/perception.png}~\textbf{Perception} task on \eventdrive \textbf{M3ED} \textbf{hard} subset. We report per-category results for \textbf{Scene Type}, \textbf{Visibility}, \textbf{Traffic Flow}, \textbf{Weather}, \textbf{Traffic Light}, and \textbf{Road Condition}, respectively.
}
\resizebox{\linewidth}{!}{
\begin{tabular}{r|cccccc}
\toprule
\textbf{Method} & \textbf{Scene Type} & \textbf{Visibility} & \textbf{Traffic Flow} & \textbf{Weather} & \textbf{Traffic Light} & \textbf{Road Condition}
\\
\midrule\midrule
\rowcolor{event_green!15}\multicolumn{7}{c}{\textcolor{event_green}{\textbf{Event-based Models}}} 
\\
EventGPT-7B \cite{liu2024eventgpt} & $69.75$ & $33.30$ & $61.82$ & $47.66$ & $52.69$ & $64.78$   \\
EventDrive-VLM                  & $78.45$ & $43.24$ & $75.08$ & $58.73$ & $65.21$ & $76.61$   \\
\midrule
\rowcolor{frame_red!15}\multicolumn{7}{c}{\textcolor{frame_red}{\textbf{Frame-based Models}}} 
\\
LLaVA-v1.6-Mistral-7B-hf \cite{llavamistral} & $90.02$ & $44.82$ & $57.05$ & $47.23$ & $52.77$ & $77.40$
\\
LLaVA-OneVision-1.5-8B \cite{llavaonevision} & $98.05$ & $65.63$ & $88.78$ & $74.05$ & $87.76$ & $97.90$
\\
InternVL2.5-8B \cite{chen2024expanding} & $96.23$ & $62.47$ & $94.78$ & $90.01$ & $89.82$ & $84.36$
\\
InternVL3-8B \cite{zhu2025internvl3} & $97.43$ & $62.82$ & $74.12$ & $82.77$ & $88.23$ & $99.14$
\\
Qwen2.5-VL-7B-Instruct \cite{qwen2.5-VL} & $95.64$ & $57.21$ & $81.14$ & $52.46$ & $85.97$ & $90.41$
\\
Qwen2.5-VL-7B-Instruct* \cite{qwen2.5-VL} & $96.18$ & $69.42$ & $88.39$ & $58.13$ & $92.44$ & $96.57$  
\\
\midrule
\rowcolor{fusion_blue!15}\multicolumn{7}{c}{\textcolor{fusion_blue}{\textbf{Event + Frame Models}}} 
\\
EventDrive-VLM   & $99.52$ & $73.36$ & $91.18$ & $59.69$ & $94.87$ & $97.93$   \\
\bottomrule
\end{tabular}
}
\label{tab:supp_m3ed_hard_split}
\end{table*}

\begin{table*}[t]
\centering
\caption{
Per-category comparison in \includegraphics[width=0.022\linewidth]{icons/perception.png}~\textbf{Perception} task on \eventdrive \textbf{PKU} \textbf{hard} subset. We report per-category results for \textbf{Scene Type}, \textbf{Visibility}, \textbf{Traffic Flow}, \textbf{Weather}, \textbf{Traffic Light}, and \textbf{Road Condition}, respectively.
}
\resizebox{\linewidth}{!}{
\begin{tabular}{r|cccccc}
\toprule
\textbf{Method} & \textbf{Scene Type} & \textbf{Visibility} & \textbf{Traffic Flow} & \textbf{Weather} & \textbf{Traffic Light} & \textbf{Road Condition}
\\
\midrule\midrule
\rowcolor{event_green!15}\multicolumn{7}{c}{\textcolor{event_green}{\textbf{Event-based Models}}} 
\\
EventGPT-7B \cite{liu2024eventgpt} & $52.62$ & $28.07$ & $67.62$ & $39.77$ & $50.28$ & $78.62$   \\
EventDrive-VLM  & $65.40$ & $42.95$ & $78.70$ & $55.47$ & $66.96$ & $92.10$   \\
\midrule
\rowcolor{frame_red!15}\multicolumn{7}{c}{\textcolor{frame_red}{\textbf{Frame-based Models}}} 
\\
LLaVA-v1.6-Mistral-7B-hf \cite{llavamistral} & $53.94$ & $28.18$ & $66.36$ & $40.30$ & $43.33$ & $70.61$
\\
LLaVA-OneVision-1.5-8B \cite{llavaonevision} & $80.30$ & $52.42$ & $90.61$ & $63.94$ & $73.03$ & $97.27$
\\
InternVL2.5-8B \cite{chen2024expanding} & $87.88$ & $52.73$ & $92.73$ & $69.39$ & $86.36$ & $91.82$
\\
InternVL3-8B \cite{zhu2025internvl3} & $79.09$ & $55.15$ & $83.64$ & $53.33$ & $86.97$ & $95.45$
\\
Qwen2.5-VL-7B-Instruct \cite{qwen2.5-VL} & $51.21$ & $51.52$ & $78.79$ & $25.45$ & $84.24$ & $73.33$
\\
Qwen2.5-VL-7B-Instruct* \cite{qwen2.5-VL} & $63.52$ & $61.43$ & $89.47$ & $34.84$ & $94.03$ & $69.98$  
\\
\midrule
\rowcolor{fusion_blue!15}\multicolumn{7}{c}{\textcolor{fusion_blue}{\textbf{Event + Frame Models}}} 
\\
EventDrive-VLM  & $71.56$ & $73.62$ & $93.45$ & $52.44$ & $95.24$ & $75.32$  \\
\bottomrule
\end{tabular}
}
\label{tab:supp_pku_hard_split}
\end{table*}

\noindent \textbf{\perception Per-dataset Results.} A cross-dataset comparison highlights the complementary strengths of each source domain and further demonstrates the robustness of \eventdrive-VLM, as shown in \cref{tab:supp_perception_dsec,tab:supp_perception_m3ed,tab:supp_perception_pku},  and \cref{tab:supp_dsec_hard_split,tab:supp_pku_hard_split,tab:supp_m3ed_hard_split}.

DSEC exhibits the highest illumination variation, including tunnels, nighttime sequences, and rapid exposure changes. Frame-based VLMs perform well on bright daytime scenes but drop significantly in low-visibility categories. Event-only models capture high-speed motion but fail on semantic cues requiring RGB textures. Our method closes this gap: it achieves strong and stable performance across all categories, demonstrating resilience to both dark scenes and motion blur.

M3ED, collected with a higher-resolution sensor suite, contains fast egomotion, extreme rotations, and diverse weather/visibility shifts across day/night conditions. Frame-only models excel when the RGB signal is clean, but degrade on nighttime or glare-heavy scenes, as seen in their instability on Visibility. Event-only models handle motion but fail on semantics. In contrast, our method consistently provides top accuracy across all categories, where motion cues and structural cues must be jointly exploited.

PKU-DAVIS-SOD, with much lower spatial resolution and relatively noisy RGB frames, is the most challenging dataset for all frame-based methods. While event-only models show some robustness in low-light scenes, their semantic performance remains limited.
Our event–frame fusion model again yields the best balanced results, showing that event signals substantially mitigate the resolution limitations of PKU’s RGB modality.

Overall, results across the three datasets indicate that neither modality alone is sufficient in diverse real-world driving scenarios. Event–frame fusion, when performed with structured alignment as in our method, offers significantly improved generality and stability across varied environments.

\noindent \textbf{\understanding Per-category Results.}
The per-category results in \cref{tab:supp_understanding} and \cref{tab:supp_understanding_hard_split} reveal clear and consistent trends across the five object-level understanding attributes.
Overall, purely event-based models struggle with fine-grained semantic cues such as Appearance, Status, and Relation-to-Others, where texture, color, and spatial detail are critical. EventGPT-7B, for instance, achieves low results on Status and Relation-to-Others, confirming that asynchronous events alone are insufficient for reasoning about detailed object properties or multi-agent relationships.

Frame-based VLMs show significantly stronger performance on appearance-heavy categories. Large models such as InternVL2.5-8B and InternVL3-8B exceed $80\%$ on Appearance, benefiting from high-resolution RGB information. However, their performance decreases on categories requiring temporal awareness and spatial consistency under motion (e.g., Object Awareness, Relation-to-Viewer). This drop is most notable for LLaVA-v1.6-Mistral, which reflects the limitations of frame-only perception under high-speed or blurred scenarios typical in DSEC.

Our \eventdrive-VLM achieves the strongest and most balanced results across all five categories. The fusion of high-frequency event cues and rich RGB semantics delivers substantial gains over both event-only and frame-only baselines. Notably, on Object Awareness, the model outperforms the best pure-frame model by a large margin. It also provides notable boosts on temporal–geometric categories (Status, Relation-to-Others), where event signals help disambiguate subtle motion cues, occlusion patterns, and relative positioning.

Although frame-based models remain competitive on highly appearance-driven categories, our method matches or surpasses them while simultaneously providing superior robustness on motion-critical attributes. These results highlight the necessity of integrating both temporal and spatial modalities for comprehensive object-level understanding in real-world driving scenes.

\noindent\textbf{Hard split Results.}
The Hard split in \cref{tab:eventdrive_hard} represents the challenging subsets in \eventdrive, characterized by extremely low illumination, aggressive high-speed motion, heavy occlusions, and substantial appearance degradation. While performance differences relative to the standard split vary across tasks and models, the Hard results provide a clearer picture of robustness under adverse conditions.

For \textbf{perception}, most models maintain relatively stable accuracy across the Hard subsets, especially on datasets where global scene semantics remain largely intact despite darker illumination. Frame-based VLMs show modest degradation on night-heavy sequences. The fused method achieves the highest scores, suggesting that combining event edges with frame semantics yields more consistent predictions under challenging visibility.

In \textbf{understanding}, the performance gap becomes more pronounced. Tasks involving grounding (\texttt{Acc@60}) and spatial consistency (\textbf{mIoU}) are especially sensitive to blur, glare, and sparse textures, leading to noticeable drops for most frame-based models. Event-driven cues mitigate some of this degradation, as high-temporal-resolution edges remain informative even when RGB details fade. The fused model again provides the most stable results, improving both grounding and mIoU by leveraging complementary spatiotemporal signals.

For \textbf{prediction}, Hard scenes introduce irregular target motion and heavy occlusions, which challenge both event- and frame-based models. Here, event cues contribute positively to short-term motion reasoning: event-only and fused models obtain higher \texttt{Speed} accuracy than most RGB-based VLMs. Path intent also benefits from the temporal consistency present in event streams, leading the fused model to reach top performance. Although the improvements are not universally large, the trend shows that motion-asynchronous signals help stabilize intent inference.

In \textbf{planning}, the difficulty of the Hard split is more evident. Nighttime ego-motion, complex turns, and GPS-denied tunnel segments create additional uncertainty for long-horizon forecasting. Event-driven temporal cues provide a more stable basis for predicting ego-vehicle dynamics, especially for acceleration and turning behavior. The fused model achieves the best results across all metrics, reducing L2 error and improving intent consistency without relying solely on appearance cues.

Across all four tasks, Hard split results highlight several general trends:

\begin{itemize}
    \item Event-only models tend to be more stable than frame-only models in conditions dominated by low light, motion blur, or rapid dynamics.
    \item Frame-based models perform strongly on standard splits but may show larger variance in grounding and spatial reasoning under degraded visibility.
    \item The fused \eventdrive-VLM consistently provides the most balanced and robust performance, benefiting from complementary strengths of event and RGB modalities.
\end{itemize}

The Hard split amplifies the value of fine-grained temporal cues, demonstrating that event streams meaningfully contribute to robustness, especially for understanding and motion-centric tasks. Overall, the Hard evaluation confirms that event signals are not merely auxiliary but play a meaningful role in maintaining reliable perception and decision-making under adverse, real-world conditions.

\section{Broader Impact \& Limitations}

\subsection{Broader Impact}
\eventdrive introduces the first full-stack event–frame multimodal benchmark for driving perception, understanding, prediction, and planning. By unifying asynchronous event streams with RGB images and language supervision, our framework pushes event-based research beyond low-level sensing toward higher-level reasoning, decision making, and explainability.

Event cameras provide microsecond temporal fidelity and high dynamic range, enabling robust perception under motion blur, low light, or extreme illumination, where conventional sensors degrade. EventDrive therefore has the potential to support safer autonomous navigation and more reliable robotic systems in challenging real-world environments.
Furthermore, EventDrive demonstrates how language grounding can be systematically incorporated into event-driven pipelines, offering a path toward interactive driving agents capable of interpreting, explaining, and justifying their decisions.

We expect that the proposed dataset and pre-training tasks will stimulate research in temporally aligned multimodal fusion, reasoning under high-frequency signals, and efficient multimodal large models. These insights may extend beyond autonomous driving to domains such as mobile robotics, AR/VR, and high-speed industrial automation.

\subsection{Societal Influence}

\eventdrive contributes to the broader vision–language community by providing a new resource that emphasizes temporal precision, sensor efficiency, and structured multimodal reasoning. Event-based vision has inherent advantages: low latency, low power consumption, and resilience to high-speed motion, which align with long-term goals of sustainable and dependable AI deployment.

From a societal perspective, improved multimodal understanding can help autonomous systems behave more predictably and transparently, promoting trust in safety-critical applications. EventDrive is curated to avoid identifiable biometric information and focuses on object-level and scene-level semantics rather than personal identity. All annotations are task-oriented and non-sensitive, targeting driving behavior analysis rather than surveillance or personal profiling.

We believe this dataset will support community-driven progress in designing interpretable, robust, and efficient event-based learning systems, while also encouraging responsible use of multimodal sensor data.

\subsection{Potential Limitations}

Despite its breadth and contributions, \eventdrive has several limitations:

\begin{itemize}
    \item \textbf{Geographic and sensor constraints.}
    The dataset covers diverse driving sequences but still originates from a limited set of cities and camera configurations. This may introduce distributional bias and reduce transferability to unseen regions, weather patterns, or sensor setups.

    \item \textbf{Automatic annotation dependency.}
    While we enforce strict validation and consistency checks, many annotations are derived using model-assisted pipelines. These may inherit biases from the underlying VLMs or from rule-based heuristics, particularly in high-level planning intent or motion interpretation.

    \item \textbf{Focus on 2D grounding and ego-centric reasoning.}
    The current benchmark emphasizes 2D bounding boxes, event–frame fusion, and ego-centric coordinate systems. World-coordinate 3D grounding and multi-agent joint planning remain out of scope, though they are natural extensions.
    
\end{itemize}
These limitations highlight opportunities for future expansion, for example, incorporating world-level 3D grounding, extending geographic coverage, or exploring ultra-long horizon reasoning.

\subsection{Ethical Considerations}

\eventdrive is designed with careful attention to ethical use and responsible data handling. All source data are from existing open-sourced datasets. All recordings are captured in public roadway environments where individuals are not identifiable, and no biometric, demographic, or personally sensitive attributes are annotated or inferable. The dataset focuses strictly on object-level and scene-level semantics relevant to autonomous driving, such as vehicles, road layout, and motion cues, rather than human identity or personal behavior patterns.

We explicitly avoid annotating or enabling tasks that could lead to privacy-invasive applications such as facial recognition, pedestrian re-identification, or profiling. Language annotations are task-specific and do not describe individuals beyond generic categories required for driving safety (e.g., “pedestrian,” “cyclist”).
Furthermore, any language generation that involves LLMs is manually filtered to ensure that no harmful or inappropriate content is introduced into the dataset.

Despite these safeguards, autonomous driving remains a safety-critical domain. Models trained on EventDrive should not be deployed directly in real-world systems without appropriate verification, robustness testing, and compliance with regional safety regulations. The work is intended solely for research purposes, and any downstream use in commercial or operational systems should incorporate additional validation and ethical review.

\section{Public Resources Used}

In this section, we acknowledge the use of the following public resources during the course of this work.

\subsection{Public Datasets Used}
We acknowledge the use of the following public datasets during the course of this work:
\begin{itemize}
    \item M3ED\footnote{\url{https://m3ed.io}.}\dotfill CC BY-SA 4.0

    \item DSEC\footnote{\url{https://dsec.ifi.uzh.ch}.} \dotfill CC BY-SA 4.0 License

    \item PKU-DAVIS-SOD\footnote{\url{https://git.openi.org.cn/LiDianze/PKU-DAVIS-SOD}} \dotfill Unknown

\end{itemize}

\subsection{Public Implementations Used}
\begin{itemize}
    \item EventGPT\footnote{\url{https://github.com/XduSyL/EventGPT}.} \dotfill Apache License 2.0
    
    \item RVT\footnote{\url{https://github.com/uzh-rpg/RVT}.} \dotfill MIT License

    \item Qwen2.5-VL-7B-Instruct\footnote{\url{https://huggingface.co/Qwen/Qwen2.5-VL-7B-Instruct}.} \dotfill Apache License 2.0

    \item InternVL3-8B\footnote{\url{https://huggingface.co/OpenGVLab/InternVL3-8B}.} \dotfill Apache License 2.0

    \item InternVL2\_5-8B\footnote{\url{https://huggingface.co/OpenGVLab/InternVL2_5-8B}.} \dotfill MIT License

    \item LLaVA-OneVision-1.5-8B-Instruct\footnote{\url{https://huggingface.co/lmms-lab/LLaVA-OneVision-1.5-8B-Instruct}.} \dotfill Apache License 2.0

    \item LLaVA-v1.6-mistral-7b-hf\footnote{\url{https://huggingface.co/llava-hf/llava-v1.6-mistral-7b-hf}.} \dotfill Apache License 2.0

     \item Impromptu-VLA\footnote{\url{https://github.com/ahydchh/Impromptu-VLA}.} \dotfill CC-BY-SA-4.0 license
    
     \item Pi3DET\footnote{\url{https://huggingface.co/datasets/Pi3DET/data}.} \dotfill MIT License

    \item PyTorch\footnote{\url{https://pytorch.org/}.} \dotfill BSD License  

\end{itemize}

\clearpage
{
    \small
    \bibliographystyle{ieeenat_fullname}
    \bibliography{main}
}

\end{document}